\documentclass[pdflatex,bst/sn-mathphys-num]{sn-jnl}

\usepackage{graphicx}%
\usepackage{multirow}%
\usepackage{amsmath,amssymb,amsfonts}%
\usepackage{amsthm}%
\usepackage{mathrsfs}%
\usepackage[title]{appendix}%
\usepackage{xcolor}%
\usepackage{textcomp}%
\usepackage{manyfoot}%
\usepackage{booktabs}%
\usepackage{algorithm}%
\usepackage{algorithmicx}%
\usepackage{algpseudocode}%
\usepackage{listings}%
\usepackage{siunitx}
\usepackage[version=4]{mhchem}
\usepackage{graphicx}
\usepackage{subcaption}
\usepackage{dsfont}

\newcommand{\ugm}{\si{\micro\gram\per\cubic\metre}}
\theoremstyle{thmstyleone}%
\theoremstyle{thmstyletwo}%

\theoremstyle{thmstylethree}%

\begin{document}

\title[Article Title]{High-resolution Nitrogen Dioxide Maps Reveal Exposure Limit Breaches across Europe}



\author*[1]{\fnm{Linus} \sur{Scheibenreif}}\email{lscheibenrei@ethz.ch}

\author[1]{\fnm{Konrad} \sur{Schindler}}\email{schindler@ethz.ch}


\affil[1]{\orgdiv{Photogrammetry \& Remote Sensing}, \orgname{ETH Zurich}, \orgaddress{\street{Stefano-Franscini-Platz 5}, \city{Zurich}, \postcode{8093}, \country{Switzerland}}}




\maketitle

\section*{Summary}
Nitrogen dioxide (\ce{NO2}) is a common air pollutant, released into the atmosphere through the incomplete burning of fossil fuels, and associated with respiratory and cardiovascular diseases in humans. Ambient \ce{NO2} concentrations are regulated through air-quality limits assessed with a sparse network of fixed monitors~\cite{eea_air_quality_reporting}. The revised EU Ambient Air Quality Directive (2024/2881) introduces a daily \ce{NO2} limit to be met from 2030~\cite{EU2024_2881}. 
At present, neither the regulatory monitoring network nor existing coarse, annual-mean models can resolve \ce{NO2} concentrations at the spatio-temporal resolutions necessary to assess compliance~\cite{cooper2020inferring, larkin2017global, shen2022europe}. Here we map \ce{NO2} across Europe at hourly and \SI{10}{\metre} resolution with a machine-learning model that combines ground monitors with satellite, reanalysis, land-use, traffic and emission data and returns a calibrated predictive distribution at every location. Validated against held-out regulatory monitors~\cite{eea_air_quality_reporting} and independent citizen-science campaigns~\cite{de2019curieuzeneuzen, lauriks2022curieuzenair}, the maps resolve high-resolution spatiotemporal \ce{NO2} gradients for 110 metropolitan areas in Europe. We reconstruct the daily compliance statistic across those regions and find limit breaches in 91 EU air quality zones deemed compliant by the regulatory monitoring network, covering a population of approximately 135\,M. Beyond air quality zones and monitor locations, an estimated 9-\num{9.4}\% (20\,M) of the population in mapped regions lives in areas where the daily \ce{NO2} limit is breached.
The high-resolution maps offer a route to population-scale assessment of compliance with the 2030 limits.

\section{Main}\label{sec:intro}
Nitrogen dioxide (\ce{NO2}) is a common air pollutant to which hundreds of millions of people are exposed every day. It forms through the incomplete burning of fossil fuels, at the exhaust of road vehicles and from power generation, industry and shipping~\cite{world2021global}. Short-term exposure inflames the airways while sustained exposure is associated with impaired lung development in children, cardiovascular disease and premature death~\cite{atkinson2018long, nitschke1999respiratory, gillespie2011respiratory, khaniabadi2017exposure}. Additionally, \ce{NO2} tracks the broader mixture of traffic-related pollutants that carry much of the health burden that can be attributed to air-pollution~\cite{beckerman2008correlation}. Because these effects begin at concentrations common in European cities, the World Health Organization sharply lowered its recommended \ce{NO2} guideline in 2021~\cite{world2021global}.

Managing the risks of air pollution depends on knowing where and when they occur, while \ce{NO2} concentration is highly variable across space and follows distinct temporal patterns at seasonal, weekly and diurnal frequencies~\cite{lebret2000small, dominguez2014spatial}. Its concentration can drop by tens of \ugm{} over a few tens of metres and drastically varies between rush hour and clear roads. Personal exposure is therefore determined by the precise locations where people live, work and travel~\cite{monn2001exposure}. Regulatory assessment, however, rests on a sparse network of fixed reference monitors~\cite{eea_aq_download_service}, and compliance is assessed in administrative air quality zones~\cite{EU2024_2881}, from a limited number of representative sampling points. The air most people actually breathe, and whether it meets the legal limits, remains largely unmeasured.

The revised European Union (EU) Ambient Air Quality Directive (2024/2881)~\cite{EU2024_2881} moves European standards towards the 2021 WHO guidance, tightening the \ce{NO2} limits that must be met from 2030 on. Alongside the existing hourly and annual limits, it introduces a daily limit. The 24-hour mean may exceed 50\,\ugm{} on no more than 18 days per year. Unlike the long-standing annual and one-hour limits, compliance with the daily limit has not been mapped or modeled at continental scale. Moreover, it increasingly constitutes the limit that is hardest to meet: as ambient \ce{NO2} has fallen across Europe over the past decades~\cite{geddes2015long}, the annual limit currently in force (40\,\ugm{}, to be lowered to 20\,\ugm{} from 2030) and the one-hour limit (200\,\ugm{}) are not exceeded at a large majority of reference stations. We find that locations that breach the daily exposure limit are frequently compliant with the annual and hourly limits.
The EU Directive identifies air-quality \emph{modeling} as a complementary means of locating non-compliance~\cite{EU2024_2881}, yet existing products are either spatially too coarse to resolve within-city gradients that determine exposure~\cite{cooper2022global, geddes2015long, cooper2020inferring}, or temporally too coarse (typically annual or monthly means) to reconstruct daily statistics~\cite{larkin2017global, shen2022europe, scheibenreif2022toward}.

Assessing where the daily limit is breached therefore demands estimates that are both fine-grained enough to capture street-scale gradients and temporally complete enough to reconstruct a daily mean for each day of the year. Here we present \ce{NO2} maps at hourly resolution and \SI{10}{\metre} ground sampling distance (GSD) across Europe. We use machine learning to fuse the observations of the regulatory network with satellite observations, atmospheric reanalysis and fine-scale land-use, traffic and emission data, to produce spatially continuous, uncertainty-calibrated maps of \ce{NO2} concentration. Based on these maps, we assess where the new daily limit is exceeded. At the level of regulatory air quality zones, we identify \ce{NO2} concentrations above the limit in 161 zones, 91 of which are deemed compliant by the current regulatory network.


\subsection{\ce{NO2} modeling approach}
\begin{figure}
    \centering
    \includegraphics[width=1.0\linewidth]{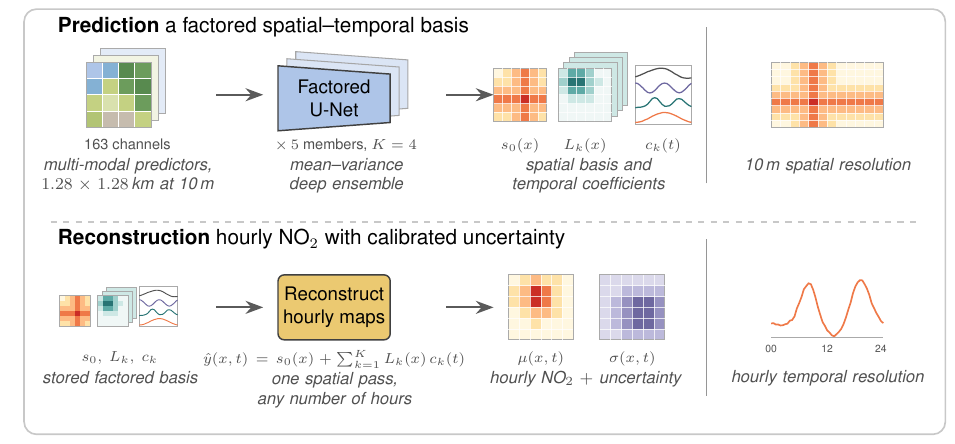}
    \caption{Overview of the \ce{NO2} estimation approach with the factored U-net model. \textbf{Top:} The model generates a spatial basis and temporal coefficients from the static and dynamic input data, respectively. \textbf{Bottom:} The spatial basis and temporal coefficients are combined to reconstruct hourly \ce{NO2} and uncertainty fields at \SI{10}{\metre} resolution according to Eqn.~\ref{eq:factored}.}
    \label{fig:model_overview}
\end{figure}

Ambient \ce{NO2} concentrations across Europe are modeled based on the \ce{NO2} record of the European Environment Agency’s air-quality monitoring network~\cite{eea_air_quality_reporting} for the years 2018–2023. We cross-reference a dataset of \num{152}\,M station-hours with a diverse set of covariates that contain determinants of \ce{NO2} pollution such as road networks, buildings, terrain and population data as well as meteorological and atmospheric reanalysis data (see Tab.~\ref{tab:channels} for the full list of covariates). We frame \ce{NO2} modelling as a sparsely-supervised dense regression problem and propose a deep ensemble model to estimate \ce{NO2} concentrations from the multi-modal input data stack. In order to assess compliance with the EU's daily \ce{NO2} limit, high-resolution estimates are required. We therefore model \ce{NO2} at hourly temporal and \SI{10}{\metre} spatial resolution. In this setting, the compute cost to cover large regions over annual time-spans ($365\times24$ predictions per pixel per year) quickly becomes prohibitive. To mitigate it, our factored U-net model processes data in a patch-wise fashion and, unlike established pixel-wise approaches~\cite{kim2021importance}, produces \ce{NO2} estimates for $1.28\times1.28$\,km areas at a time, at \SI{10}{\metre} resolution. Additionally, the \ce{NO2} field is factorised into static spatial bases and temporally varying coefficients, which amortises the expensive processing of high-resolution spatial inputs over a full year of hourly temporal coefficients (Fig.~\ref{fig:model_overview}, see Sec.~\ref{sec:modeling} for details). The model is trained on $\approx85$\% of the monitoring locations (127\,M station-hours) and evaluated on the remaining 15\% of locations (25\,M station-hours), where it outperforms pixel-wise and reanalysis-based baselines (Tab.~\ref{tab:results_heldout_monitors}) and reaches an RMSE of $11.32\pm0.26$\,\ugm{} and R2-score of $0.571\pm0.013$. The model accurately reconstructs temporal patterns of \ce{NO2} concentration across different time scales (Figs.~\ref{fig:temporal_cycles},~\ref{fig:by_year},~\ref{fig:madrid_hourly}) and corresponds well to high-resolution spatial patterns established in independent citizen science campaigns~\cite{de2019curieuzeneuzen, lauriks2022curieuzenair} in Flanders 2018 (RMSE $4.25$\,\ugm{}, R2-score $0.562$) and Brussels 2021 (RMSE $5.05$\,\ugm{}, R2-score $0.470$).
To facilitate interpretation of the model's \ce{NO2} estimates, we also provide estimates of both aleatoric uncertainty (obtained through training with a Gaussian log-likelihood loss~\cite{nix1994estimating}) and epistemic uncertainty (obtained from the spread between ensemble members~\cite{lakshminarayanan2017simple}). Together, they specify a calibrated predictive distribution per pixel (Fig.~\ref{fig:uncertainty_calibration}, see Sec.~\ref{sec:uncertainty}), which we utilise for a probabilistic assessment of where the mapped values exceed the regulatory limit.

\subsection{Limit breach detection}
The EU's daily \ce{NO2} limit is breached if, within a year, more than 18 days are recorded with a 24-hour mean \ce{NO2} concentration above 50\,\ugm{}. We assess compliance with the regulatory limit based on two statistics reconstructed from our model's full hourly \ce{NO2} timeseries.
The \emph{expected non-compliance} statistic computes the expected number of days exceeding 50\,\ugm{} from the modeled hourly predictive mean and variance.
Additionally, we derive the \emph{probability of non-compliance} as the probability of observing more than 18 days above 50\,\ugm{} based on the predictive uncertainty of the model (see Sec.~\ref{sec:limit_breach_detection} for details).

\paragraph{Air quality zones}

\begin{figure}
    \centering
    \includegraphics[width=1\linewidth]{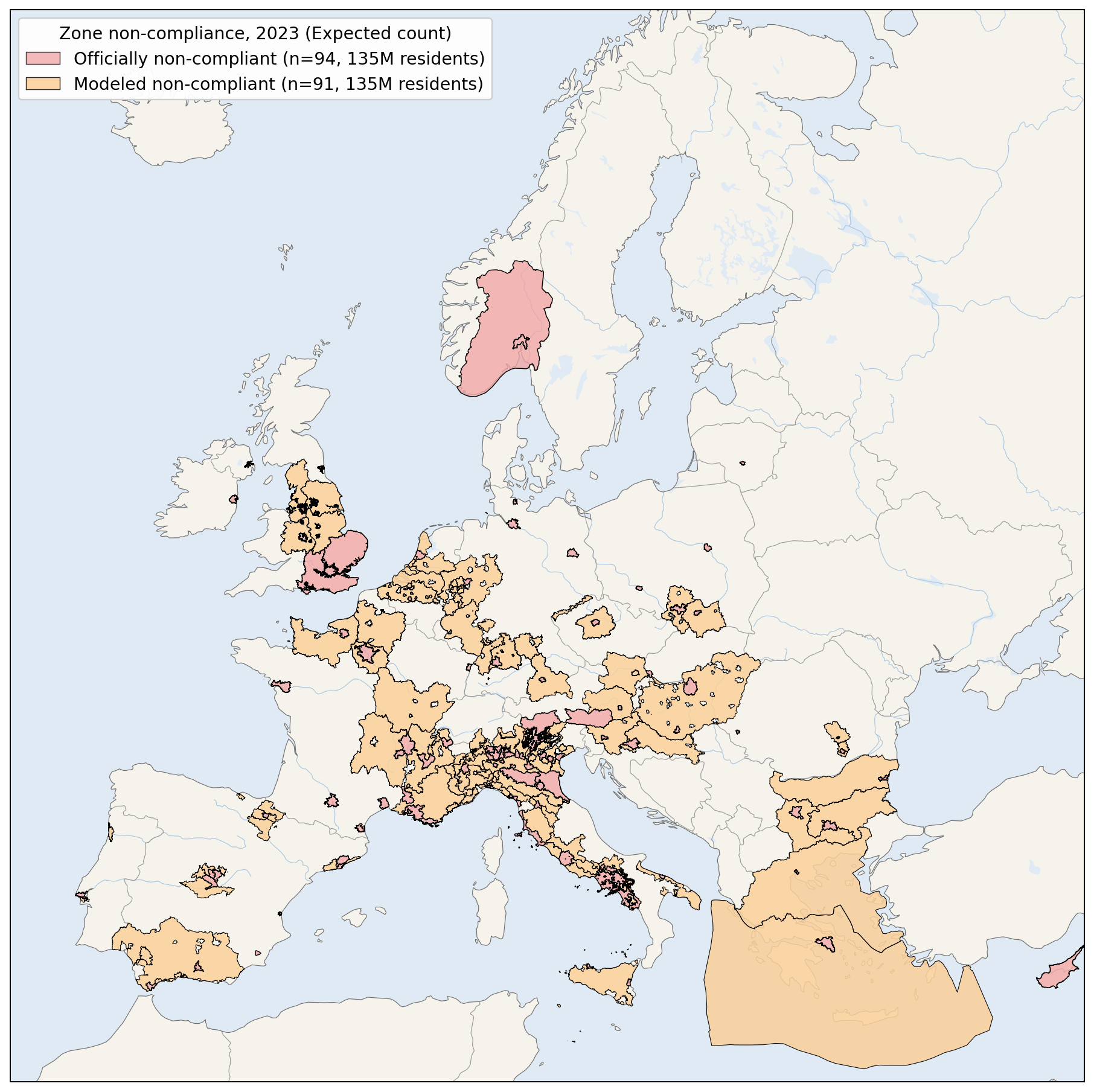}
    \caption{Overview of EU air quality zones with limit breaches detected by regulatory monitors (red) and additional detections from modeled \ce{NO2} concentrations (orange).}
    \label{fig:breaching_zones}
\end{figure}

\begin{figure}
    \centering
    \includegraphics[width=1.\linewidth]{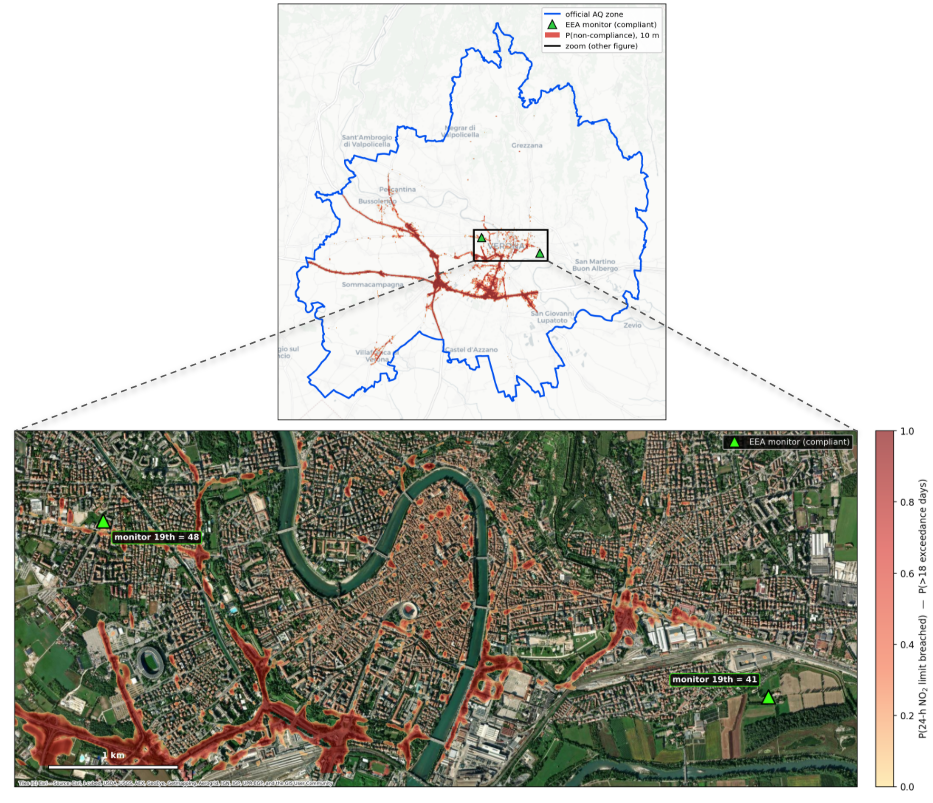}
    \caption{EU air quality zone \texttt{IT.0521} in Verona, Italy. \textbf{Top:} Map of the air quality zone with two compliant regulatory monitors in 2023. \textbf{Bottom:} Modeled limit breach probability in the vicinity of the compliant regulatory monitors in Verona.}
    \label{fig:breaching_zone_verona}
\end{figure}

\begin{figure}
    \centering
    \includegraphics[width=0.75\linewidth]{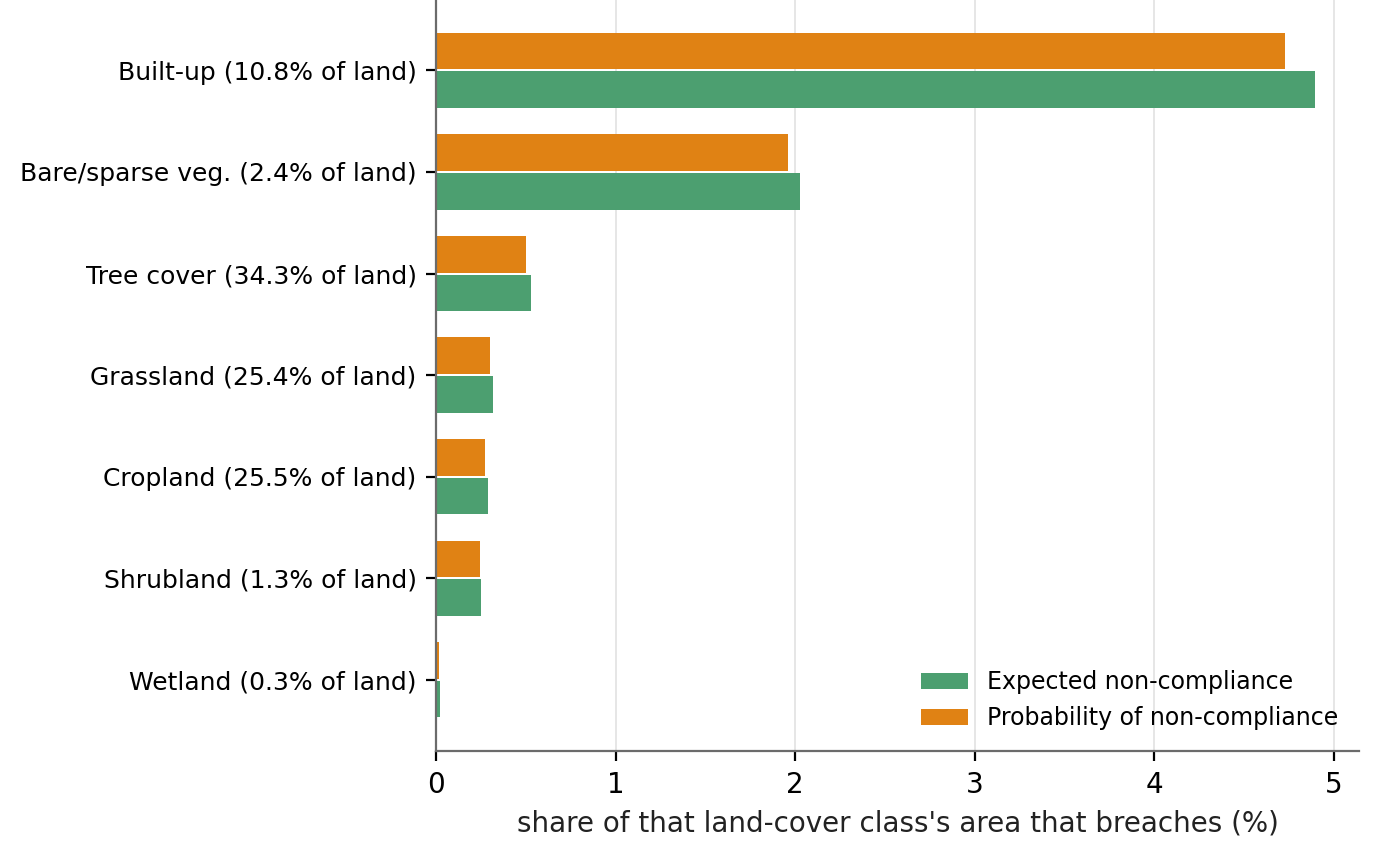}
    \caption{{\textbf{Breach rate by land-cover.} Share of limit breaching pixels by land-cover type for the expected breach statistic (green) and the probabilistic estimate (orange).}}
    \label{fig:breaches_by_landcover}
\end{figure}

The EU Ambient Air Quality Directive~\cite{EU2024_2881} delineates Europe into air quality zones, covering a total of
\num{520.8}\,M people. Of \num{3060} EEA stations reporting \ce{NO2} in 2023, \num{264} breach the daily limit. Air quality zones are deemed non-compliant with the limit if they contain at least one such station~\cite{EU2024_2881}. This yields 94
non-compliant zones across Europe, covering
\num{134.9}\,M residents. We estimate the expected number of limit exceedance days in 110 metropolitan regions (Fig.~\ref{fig:mapped_regions}) based on modeled hourly \ce{NO2} concentrations at \SI{10}{\metre} resolution and find 161-164 air quality zones and \num{252}-\num{256}\,M residents to be non-compliant (Fig.~\ref{fig:breaching_zones}, probability of non-compliance and expected non-compliance, respectively). Of the flagged zones, 73 zones agree with the regulatory network and 91 zones, home to \num{134.8}\,M residents, are identified only by our model. We identify frequent limit breaches near city-center and residential areas (Fig.~\ref{fig:breaching_zone_verona}) and close to regulatory monitors whose observations lie just below the daily limit.
Breaches predominantly occur in built-up regions, affecting 4.8\% of total built-up area, versus 0.5\% of tree-covered regions and 0.3\% of grasslands/croplands. Only 10.8\% of the mapped area are built-up, but they account for 58\% of all breaches (Fig.~\ref{fig:breaches_by_landcover}).

\begin{figure}
    \centering
    \includegraphics[width=1.\linewidth]{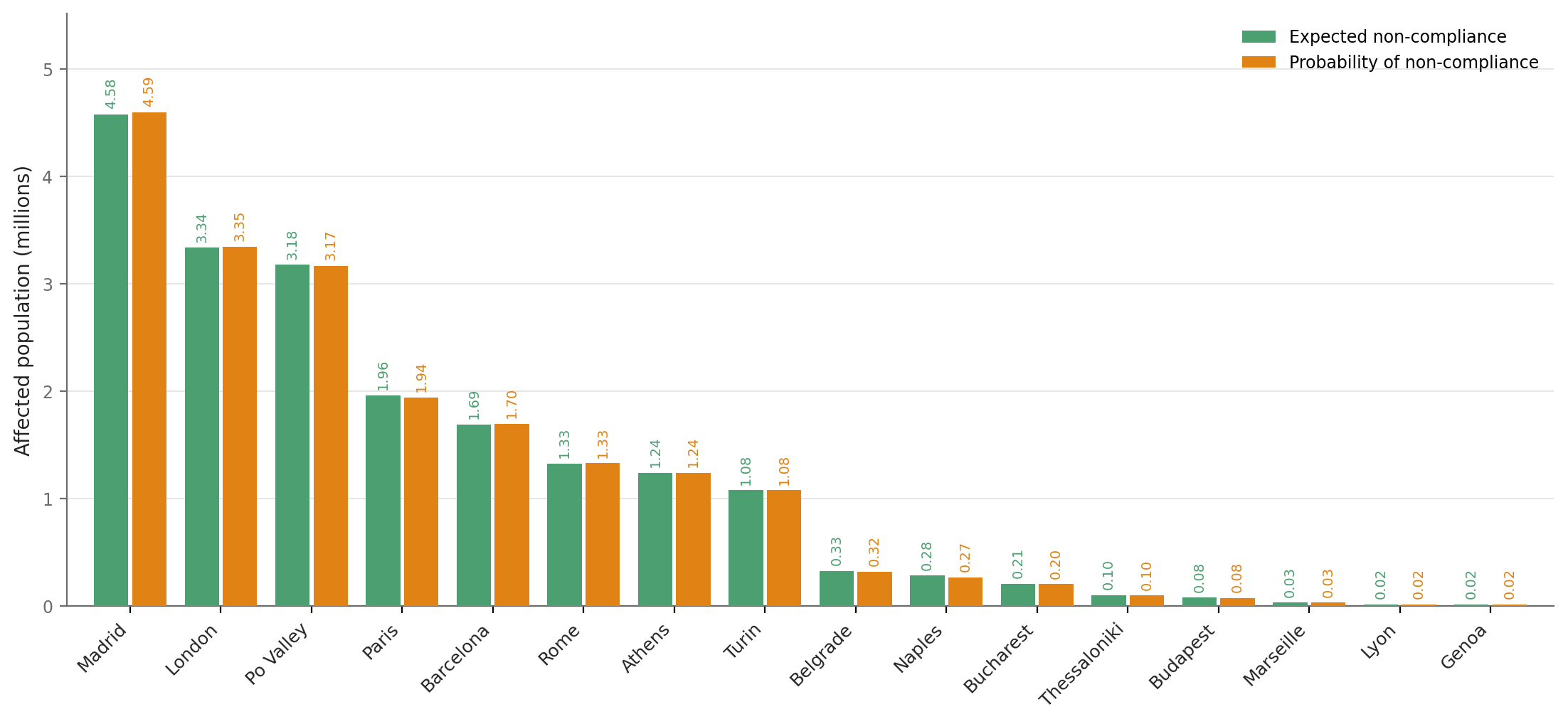}
    \caption{\textbf{Population living in breach.} Population living in breach of the daily \ce{NO2} limit in selected regions according to the expected breach statistic (green) and the probabilistic estimate (orange).}
    \label{fig:breaching_population_by_region}
\end{figure}

\paragraph{Population exposure}
The 110 mapped regions (Fig.~\ref{fig:mapped_regions}) are home to 215 M people, roughly 40\% of the EU-27(+UK) population.
Our expected exceedance statistic finds 20.15\,M people (9.4\%) affected by a 24-hour-limit breach, and 19.49\,M (9.1\%) live in areas with a probability of non-compliance greater than 50\%. Exposure shows strong geographic patterns, concentrated in southern-European conurbations (Fig.~\ref{fig:breaching_population_by_region} Madrid 4.5 M, the Po Valley 3.2 M, Barcelona, Rome, Turin, Athens) together with London and Paris, while the large northern metropolitan regions (Rhine-Ruhr, Randstad, the English Midlands, Upper Silesia) show essentially no 24-hour-limit breaches.

The compliance statistics evaluated on held-out monitors detect true limit breaches with high precision (0.892–0.910, Tab.~\ref{tab:breach-confusion}, Fig.~\ref{fig:observed_vs_predicted_19th}) but recover only 34–35\% of them. Therefore, the newly identified 91 zones (134.8\,M residents) and 20\,M people beyond those already detected by the regulatory network constitute a conservative lower bound for hidden non-compliance, not a complete accounting.  The true extent of daily \ce{NO2} limit breaches throughout Europe is almost certainly larger.

Breach days are heavily concentrated in the winter, with 54\% of observed breach-days in November–February and a clear minimum in May–August. Breach days are also not temporally independent: 78\% of observed breach-days occur in a run of $\geq 2$ consecutive days. Long runs of 6 or more days account for 26\% of all breaches. Within the mapped regions, the number of exceedance days per pixel is heavily skewed to the left, with 5.4\% of mapped pixels exceeding the limit on at least one day of the year, but only 0.41\% exceeding it on more than 18 days. Breaches are concentrated in a small set of roadside hotspot pixels that exceed the limit on 100–270 days of the year.

Overall, we find that exceedance of the 24-hour limit is not a diffuse, year-round phenomenon but tends to occur in marked, spatially and seasonally concentrated multi-day episodes, with direct implications for the health burden.

\subsection{Breach severity}
We also examined how close currently-compliant zones are to the \ce{NO2} limit, based on both the regulatory network's measurement and the model's probabilistic risk. Of the 225 officially-compliant, monitored zones in our mapped footprint, 78 (35\%) have at least one monitor whose own 19\textsuperscript{th}-highest daily mean already falls within $[40, 50)$\,\ugm{}, i.e., less than 10\ugm{} from the limit. The remaining 147 (65\%) have a larger margin to the limit, remaining below $40$\,\ugm{}.
The EU Directive treats air-quality modeling as a complementary tool for locating non-compliance. We find that the model's disagreement with the regulatory network concentrates where the official measurements are already close to the limit. Of 78 zones that are at most 10\,\ugm{} below the limit according to the official statistics, 43 (55\%) are flagged in our map as breaching the limit. This share rises to 71\% for zones that are within 5\,\ugm{} of the limit. As an illustrative example, the Verona zone (Fig.~\ref{fig:breaching_zone_verona}) has two monitors below the limit, but unmonitored locations nearby have a high breach probability under the model.

At zone level, 164 and 161 of the 670 official zones are flagged by the compliance statistics, respectively. Breaches are frequently detected on roads and if road pixels are excluded from the assessment, 30 (18.3\%) and 27 (16.8\%) zones, respectively, are instead compliant under a wide road width assumption (i.e, flip from non-compliant to compliant), and 22 (13.4\%) and 21 (13.0\%) when assuming a narrow width (Sec.~\ref{sec:road_breaches}). Roughly four fifths of non-compliant zones therefore remain non-compliant away from their roads. The zones that would change status are home to 48.5 and 44.5\,M residents for the expected and probabilistic non-compliance statistics, respectively.

\section{Methods}
\subsection{Datasets}
\paragraph{Regulatory \ce{NO2} network}
We model ambient \ce{NO2} concentrations across Europe at hourly resolution. Ground truth is the hourly \ce{NO2} record of the European Environment Agency's air-quality monitoring network~\cite{eea_air_quality_reporting} for the years 2018--2023, comprising \num{3823} reference stations spanning urban-traffic, urban-background, suburban and rural site types. Each supervised example corresponds to a single station--hour pair. We collect a dataset of \num{152}M station--hours across Europe.
The dataset is restricted to observations of hourly \ce{NO2} concentrations that are tagged as \texttt{valid} with verification by the local reporting authority. Additionally, measurements below zero and spikes above 500\,\ugm{} are not considered.

\paragraph{Land-use, satellite and reanalysis data.}
For every station-hour observation from the EEA network we collect a stack of multi-modal covariates colocated with the monitor, resampled to \SI{10}{\metre} resolution and covering an area of $1.28\times\SI{1.28}{\kilo\metre}$ around the monitor. Predictors combine temporally static land-use, emission and topographic descriptors with dynamic reanalysis and satellite data that vary at different temporal frequencies (see Tab.~\ref{tab:channels} for the full list of covariates). 

\begin{table}[t]
\centering
\caption{Covariates in the \ce{NO2} estimation dataset.}
\label{tab:channels}
\small
\begin{tabular}{@{}llll@{}}
\toprule
Modality & Dynamic & Static & Source \\
\midrule
Learned geospatial embeddings & & \checkmark & AlphaEarth foundations~\cite{brown2025alphaearth}\\
Digital elevation model & & \checkmark & Copernicus GLO-30~\cite{strobl2020new} \\
Roads, buildings, land-use & & \checkmark & Rasterized from OpenStreetMap~\cite{OpenStreetMap} \\
Population density & & \checkmark & WorldPop~\cite{tatem2017worldpop} \\
Road traffic / emissions & & \checkmark & CAMS-REG-AP v8.1~\cite{kuenen2022cams} \\
Atmospheric reanalysis & \checkmark & & CAMS regional \ce{NO2}/\ce{NO_x}~\cite{kuenen2022cams} \\
Land cover & & \checkmark & ESA WorldCover~\cite{zanaga2022esa} \\
Weather reanalysis & \checkmark & & ERA5~\cite{hersbach2020era5} \\
\ce{NO2} column density & \checkmark & & Sentinel-5P TROPOMI~\cite{veefkind2012tropomi} \\
Built-up population & & \checkmark & GHSL~\cite{schiavina2023ghs} \\
Annual average daily traffic & & \checkmark & Traffic flow estimates~\cite{shen2024europe} \\
\bottomrule
\end{tabular}
\end{table}

We distinguish static per-pixel layers that capture fine-scale determinants of the \ce{NO2} field (street canyons, the road network, buildings, terrain, population), and dynamic layers with meteorological and photochemical information (wind speed and direction, boundary layer, temperature, regional emissions and satellite \ce{NO2} observations) that modulate concentrations over time. Additionally, we derive dynamic terms such as an "is-weekend" indicator, a normalized day-of-year, a year index, geolocation encodings, an Sentinel-5P-missing mask and a 7-day Sentinel-5P backfill. We also derive additional covariates from static data sources. 
The elevation data~\cite{strobl2020new} is projected to different scales by a discrete wavelet decomposition and independent reconstruction of the coefficients to 12 scale-specific elevation maps (\SI{30}{\metre}--\SI{122.9}{\kilo\metre}, as proposed in~\cite{kim2021importance}).
We obtain vector layers from OpenStreetMap~\cite{OpenStreetMap} (roads by class, buildings, land use, water aeroway, ports) and rasterize them independently into binary maps.
AlphaEarth embeddings are dequantized according to their usage instructions~\cite{brown2025alphaearth}.
The annual-average-daily-traffic data~\cite{shen2024europe} is accumulated within nested buffer radii (\SIrange{50}{5000}{\metre}) around each pixel.
We utilize the following variables from ERA5~\cite{hersbach2020era5}: total precipitation (TP), boundary layer height (BLH), clear-sky direct solar radiation (CDIR), temperature at 2m (T2M) and wind (U10,V10).

Land-use predictors that are heavy-tailed or non-negative are transformed per-channel with a fitted Yeo--Johnson power transform~\cite{weisberg2001yeo} followed by z-score standardization (TP, BLH, CDIR, CAMS \ce{NO2} reanalysis and emissions, population density, traffic data and the target \ce{NO2} measurement). Roughly symmetric covariates are z-score standardized (T2M, U10, V10, DEM incl. wavelet levels). The geospatial embeddings, binary OSM and WorldCover maps as well as time and location encodings are left in their native scale. Normalization statistics are estimated once on the training split and applied unchanged to validation, test and inference data.

\begin{figure}
    \centering
    \includegraphics[width=1.\linewidth]{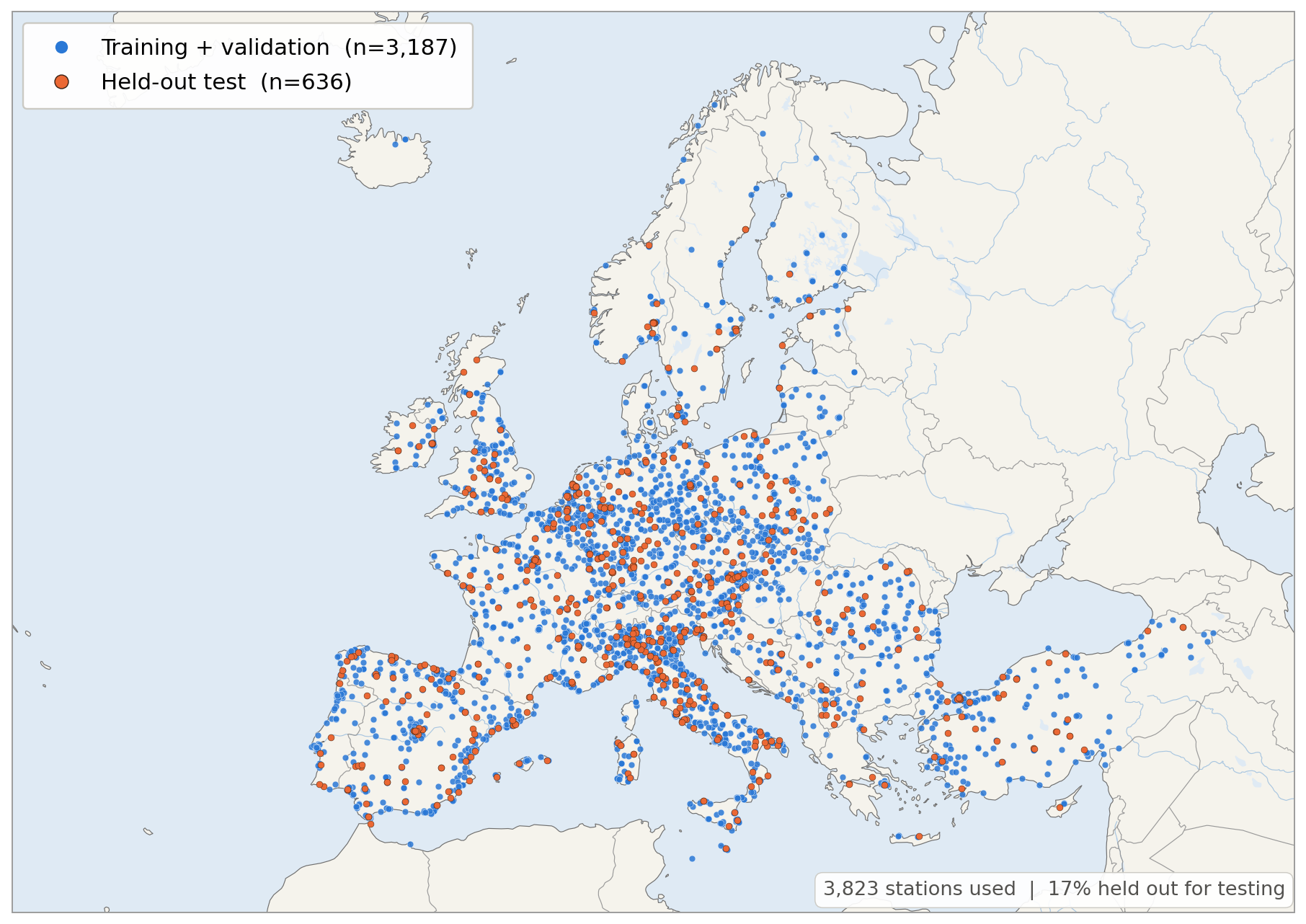}
    \caption{Overview of the regulatory \ce{NO2} monitoring network. We utilize hourly measurements for the 2018-2023 period and split the locations into training, validation and test sets.}
    \label{fig:eea_monitors_by_split}
\end{figure}

\paragraph{Dataset splits}
Monitoring stations are partitioned into distinct training, validation and test sets. Each station is described by its static context: coordinates, population, distance to the nearest major road, the DEM/wavelet terrain data, and its long-run mean observed \ce{NO2}. These features are standardized, log-transformed where skewed, reduced to five dimensions with UMAP~\cite{mcinnes2018umap} and clustered with HDBSCAN~\cite{mcinnes2017hdbscan} (following the approach by~\cite{kim2021importance}). Within each cluster the stations are drawn at random into 70\% training, 15\% validation and 15\% test, so every split carries the same mixture of station environments (Fig.~\ref{fig:eea_monitors_by_split}). In addition, we filter for locations with at least 365 days carrying \num{\geq18} hourly values together with plausible statistics (mean, standard deviation and 99\textsuperscript{th} percentile below \num{100}, \num{30} and \num{200}\,\ugm{}, respectively).

\paragraph{Citizen science \ce{NO2} Campaigns}
For independent evaluation we use \ce{NO2} datasets from passive sampler campaigns that are not part of the regulatory network. These are the citizen-science CurieuzeNeuzen campaign in Flanders from May 2018 with \num{17886} diffusion tube \ce{NO2} concentration measurements~\cite{de2019curieuzeneuzen} and the CurieuzenAir campaign in Brussels~\cite{lauriks2022curieuzenair} (25 September--23 October 2021, $\approx$2,493 tubes).
Passive samplers integrate \ce{NO2} over a multi-week window, so the corresponding model estimate is the mean prediction over the same window at the sampler location. These campaigns provide high spatial coverage relative to the regulatory network, however the geo-location of each diffusion tube is only known to street level to ensure data privacy. 

\subsection{\ce{NO2} modeling}
\label{sec:modeling}

We cast hourly \ce{NO2} mapping as a sparsely-supervised dense regression problem. A convolutional network processes the multi-modal data patch and outputs a dense \ce{NO2} field at \SI{10}{\metre} ground resolution ($1.28\times\SI{1.28}{\kilo\metre}$ per patch). The loss is evaluated only at the exact location of the supervising station, via a masked objective. Cases where multiple monitoring stations are located within the same patch are rare but explicitly handled, and every corresponding pixel is supervised. A distinct prediction is produced for every hour of the 2018--2023 record with independent inference steps. At deployment, dense maps are produced per UTM tile at \SI{10}{\metre} and hourly cadence.

\paragraph{Model architecture}
We use a factored U-net~\cite{ronneberger2015u} model architecture that consists of a three-level encoder--decoder with GroupNorm and SiLU activations. The model maps the patch of static input modalities to a time-invariant spatial baseline field $s_0(x)$ and $K$ spatial mode maps $L_k(x)$. A small temporal head (implemented as two layer MLP) maps the dynamic inputs at hour $t$ to $K$ scalar coefficients $c_k(t)$. The hourly prediction is their linear combination,
\begin{equation}
  \hat{y}(x, t) = s_0(x) + \sum_{k=1}^{K} L_k(x)\, c_k(t),
  \label{eq:factored}
\end{equation}
implemented with $K=4$ temporal modes (Fig.~\ref{fig:model_overview}). This factorization of the input dataset into static and dynamic components makes dense inference at hourly resolution feasible. The spatial encoder is evaluated only once per location and year, after which any number of hours is reconstructed from the cheap temporal head and the linear combination of temporal coefficients and spatial maps (Eqn.~\ref{eq:factored}). This reduces a full year of hourly maps from $365\times24$ forward passes to one spatial pass plus per-hour coefficients (Fig.~\ref{fig:model_overview}).

The factored U-net model has \num{2.25}M trainable parameters, including a \num{128} channel convolutional stem that maps the predictor stack to the \num{64} channel U-net base.

\paragraph{Uncertainty quantification}\label{sec:uncertainty}
The full \ce{NO2} model is a mean--variance deep ensemble~\cite{lakshminarayanan2017simple} of 5 factored U-net models, trained from independent random weight and dataset initializations. Besides the \ce{NO2} baseline and mode maps, each network additionally emits a per-pixel predictive log-variance for each of them. It is trained with Gaussian negative-log-likelihood, so that a single model produces a heteroscedastic aleatoric uncertainty~\cite{nix1994estimating}. The disagreement among ensemble members supplies the epistemic component. The two are combined into a single calibrated predictive distribution
per pixel as a Gaussian mixture, with mean $\bar\mu = \tfrac1M\sum_m \mu_m$ and variance $\bar\sigma^2 = \tfrac1M\sum_m(\sigma_m^2 + \mu_m^2) - \bar\mu^2$. Both the variance head and the ensemble spread are reconstructed through the same factored
basis, so the calibrated $\sigma$ is available densely and hourly at no additional inference cost.
The network is trained by Gaussian negative-log-likelihood in $\log(1+y)$ space with the $\frac{1}{2}\Big(\log\sigma^2 + \frac{(\mu-\log(1+y))^2}{\sigma^2}\Big)$ objective. Hence, transforming the output with $e^\mu - 1$ back into the \ce{NO2} concentration space returns the predictive median, underestimating the log-normal mean.
We therefore report the analytic predictive mean, $e^{\mu + \sigma^2/2}-1$.

\paragraph{Limit breach detection}\label{sec:limit_breach_detection}
The factored U-net model provides a predictive distribution, which we use to derive a set of compliance statistics for the daily EU \ce{NO2} limit.
For a target station and year we predict every observed station--hour, obtaining a Gaussian predictive distribution $\mathcal{N}(\mu_h, \sigma_h^2)$ per hour $h$. Hours are grouped into local-calendar days and a day $d$ is retained only if it has at least 18 valid hours $H_d$, station-years with fewer than 300 valid days are dropped. We distinguish between two probabilistic formulations of the limit breach statistic:


\textbf{Expected non-compliance.} The predicted daily mean is the average of the hourly means, $\mu_d = \frac{1}{H_d}\sum_1^{H_d}\mu_h$, and its predictive variance $\sigma_d^2$ is propagated from the constituent hourly variances, so that each day carries its own distribution $\mathcal{N}(\mu_d, \sigma_d^2)$ for the daily mean. Ranking the predicted daily means $m_d$ in descending order, the 19\textsuperscript{th}-highest daily mean is the regulatory compliance statistic. A location is flagged if this value exceeds 50\,\ugm{}.
We report the expected value of the exceedance day count based on the model's predictive distribution, $\sum_{d=1}^{D} \Phi\!(\frac{\mu_d - 50}{\sigma_d})$.

\textbf{Probability of non-compliance.} We propagate the model's predictive uncertainty into a probability of non-compliance with the daily EU \ce{NO2} limit. For each model ensemble member the predicted daily mean $\mu_{m,d}$ and variance $\mathrm{Var}_{m,d}=\frac{1}{H_d^2}\sum_{h=1}^{H_d}\sigma_{m,h}^2$ are computed from the hourly estimates. We then aggregate mean and variance across ensemble members $m\in\mathrm{M}$ as:
$$
\mu_d = \frac{1}{M}\sum_{m=1}^{M}\mu_{m,d} \qquad \sigma_d^2 = \frac{1}{M}\sum_{m=1}^M \mathrm{Var}_{m,d} + \frac{1}{M}\sum_{m=1}^M\mu_{m,d}^2-\mu_d^2
$$
The daily \ce{NO2} distribution is modeled as Gaussian with mean $\mu_d$ and variance $\sigma_d^2$, which gives a daily limit (\num{50}\,\ugm) exceedance probability of $p_d=\mathrm{P}(\bar{c}_d > 50) = \Phi\Big(\frac{\mu_d-50}{\sigma_d}\big)$. We extend this to the annual probability of non-compliance (at least 19 days of daily limit exceedances) as $\mathrm{P}(N>18)$, where $N=\sum_1^DX_d$ is the number of exceedance days in the year, each of which is Bernoulli distributed with probability $p_d$. The number of exceedance days $N$ is approximated with a Gaussian with $\mu_N=\sum_{d=1}^D p_d$ and $\sigma_N^2=\sum_{d=1}^D p_d\cdot(1-p_d)$. This gives the probability of a limit breach as:
$$
P(N > 18) \approx \Phi\Big(\frac{\mu_N - 18}{\sigma_N}\Big).
$$
In practice we use a continuity correction and consider $P(N > 18.5)$ when approximating the discrete count statistic with a continuous Gaussian.

\begin{figure}
    \centering
    \includegraphics[width=1.\linewidth]{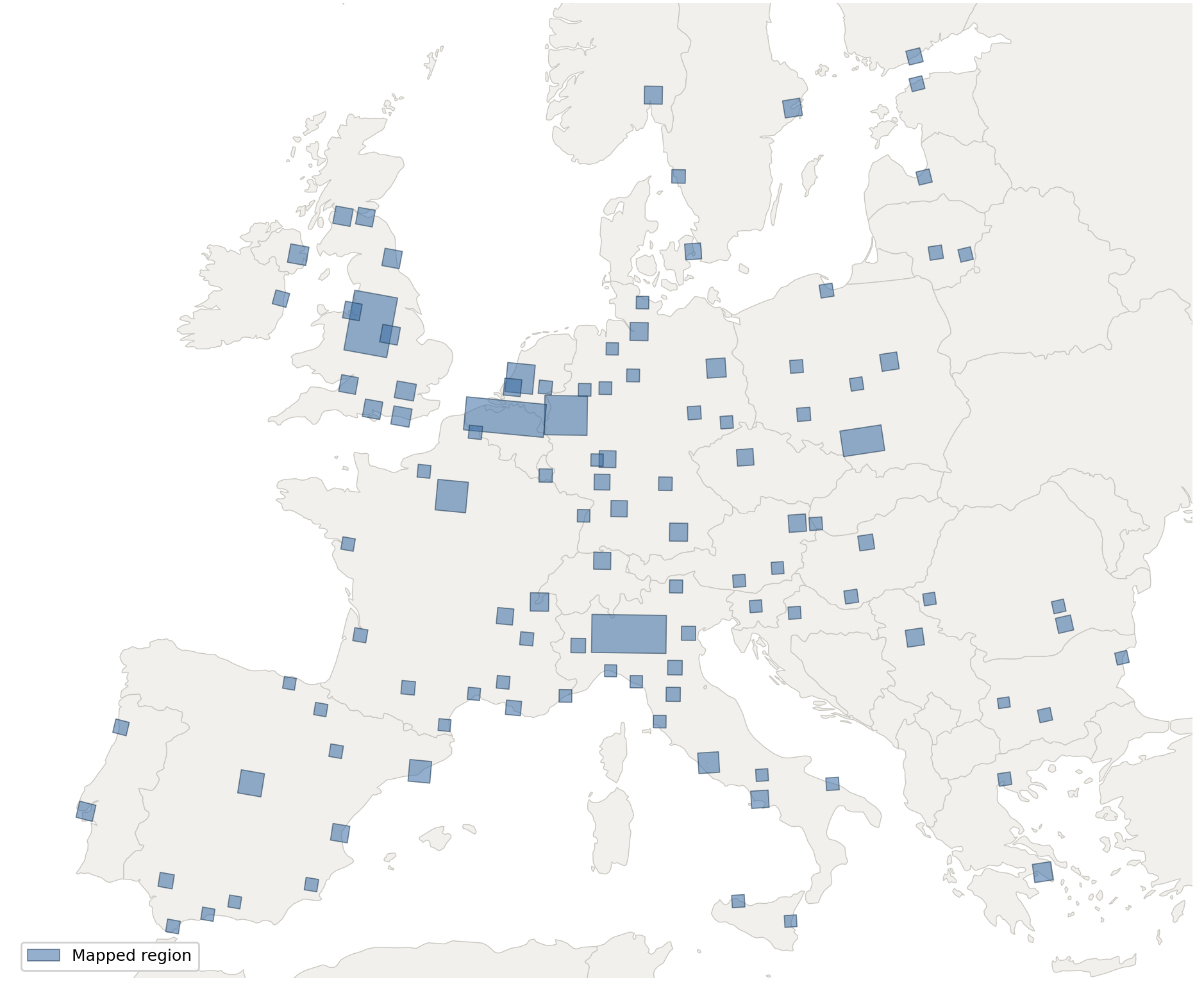}
    \caption{\textbf{Mapped regions.} Regions across Europe for which compliance with the EU daily \ce{NO2} limit was assessed at \SI{10}{\metre} resolution (n=110, see Tab.~\ref{tab:mapped_regions} for the full list).}
    \label{fig:mapped_regions}
\end{figure}

\paragraph{Dense inference}
Continuous maps over an area of interest are produced by tiling the region on its native UTM grid, evaluating the spatial encoder once per tile to obtain the factored basis ($s_0$, $L_k$, and the log-variance bases), and reconstructing each hour from per-tile-centre dynamic vectors. Per-hour fields are combined into temporal aggregates (e.g.\ annual or campaign-window means) analytically from the stored basis, and adjacent tiles are seam-blended and mosaicked together. Land-surface masking uses the WorldCover water class~\cite{zanaga2022esa}.
In total, we map compliance statistics for 110 regions across Europe in 2023 (Fig.~\ref{fig:mapped_regions}), covering all major metropolitan areas. We selected the EU-27 capital cities, metropolitan regions with over 1\,M inhabitants and added smaller cities in the vicinity of already mapped areas (see Tab.~\ref{tab:mapped_regions} for the full list). The covered areas vary in size between \num{690} (Genoa) and \num{24287}\,km\textsuperscript{2} (Po Valley). Region sizes are selected proportionally to the square-root of the covered population. Where appropriate, individual regions are combined into larger areas (Po Valley, Midlands North, Flanders, Rhine-Rhur, Upper Silesia).

\paragraph{Model evaluation}
\textbf{Held-out monitoring stations.} We evaluate the Factored U-net ensemble model on held-out regulatory stations across 2018-2023. On \num{636} independent locations with \num{25}\,M station-hours (\num{15}\,M urban, \num{6}\,M suburban, \num{4}\,M rural), the model reaches an RMSE of \num{11.32}\,\ugm{}, MAE of \num{7.21}\,\ugm{}, and R2-score of \num{0.57} (Tab.~\ref{tab:results_heldout_monitors}).
Performance is best at monitors classified \texttt{background} (RMSE \num{8.89}\,\ugm{}), but lower in \texttt{industrial} or \texttt{traffic} settings (RMSE of \num{10.24} and \num{14.53}\,\ugm{}, respectively).
\begin{table}[]
    \centering
    \caption{Evaluation of different models for hourly \ce{NO2} mapping across 636 held-out regulatory monitors, covering 25\,M station-hours in 2018-2023. Standard errors from 1000 draws of station-clustered bootstrap.}
\begin{tabular}{@{}rrrrrrrrr@{}}
\toprule
Method & RMSE & MAE & Bias & $R^{2}$ & $r$ \\
\midrule
Constant & $17.29\pm0.34$ & $12.85\pm0.18$ & $-0.00\pm0.41$ & $0.000\pm0.000$  & -- \\
Raw CAMS & $14.89\pm0.41$ & $8.88\pm0.27$ & $-6.71\pm0.31$ & $0.259\pm0.022$ & $0.642\pm0.011$ \\
CAMS OLS & $13.26\pm0.32$ & $9.07\pm0.17$ & $-0.06\pm0.31$ & $0.412\pm0.146$ & $0.642\pm0.011$ \\
Ridge regression & $11.84\pm0.27$ & $7.92\pm0.16$ & $0.09\pm0.23$ & $0.517\pm0.011$ & $0.719\pm0.008$ \\
Factored U-net & $11.32\pm0.26$ & $7.21\pm0.16$ & $-0.47\pm0.21$ & $0.571\pm0.013$ & $0.756\pm0.008$ \\
\bottomrule
\end{tabular}
    \label{tab:results_heldout_monitors}
\end{table}

\begin{figure}
    \centering
    \includegraphics[width=1.\linewidth]{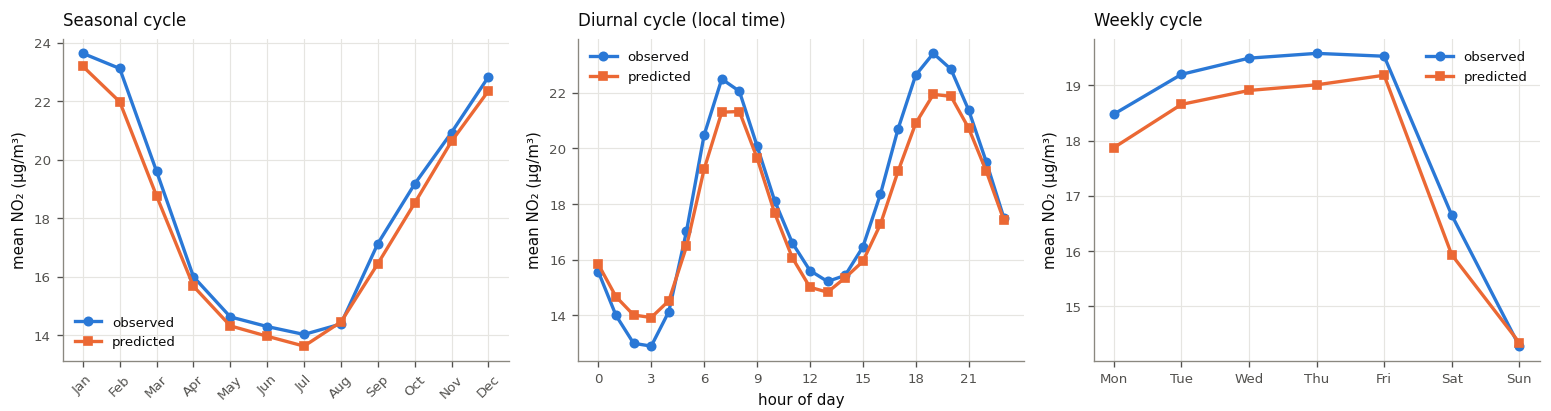}
    \caption{\textbf{Temporal \ce{NO2} trends.} Observed and predicted mean \ce{NO2} concentrations at held-out regulatory monitors across different timescales.}
    \label{fig:temporal_cycles}
\end{figure}

\begin{figure}
    \centering
    \includegraphics[width=1.\linewidth]{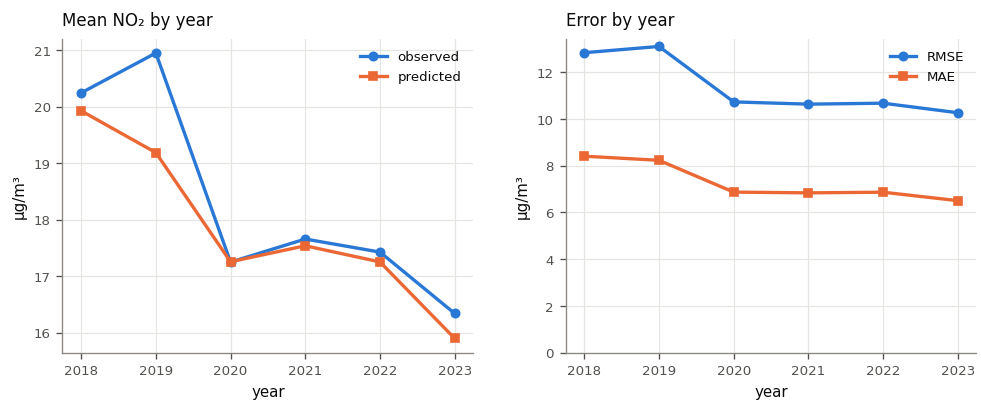}
    \caption{Observed and predicted annual mean \ce{NO2} concentrations at held-out regulatory monitors.}
    \label{fig:by_year}
\end{figure}

\begin{figure}
    \centering
    \includegraphics[width=1.\linewidth]{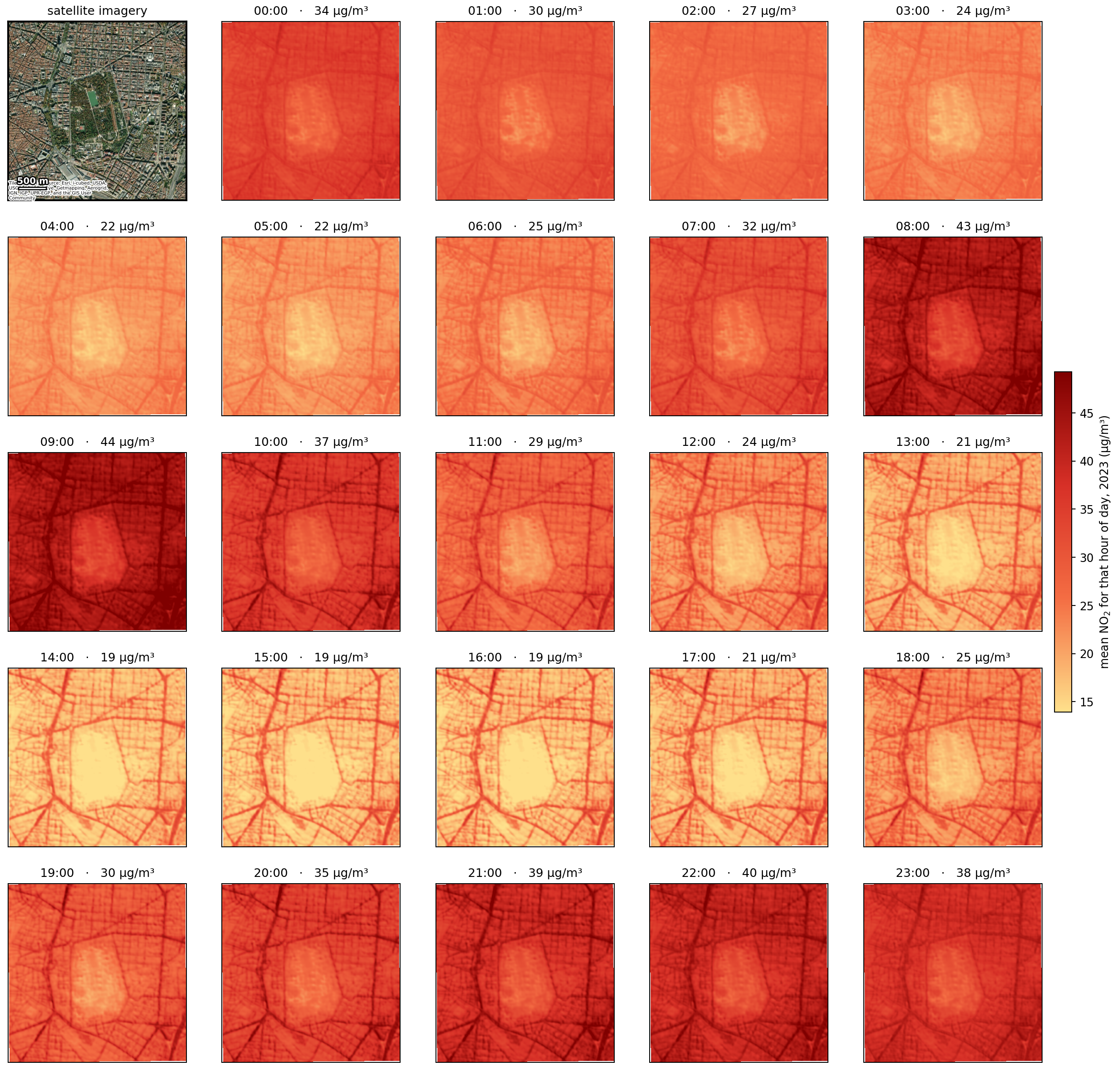}
    \caption{\textbf{Hourly \ce{NO2} maps for Madrid.} Hourly average of modeled \ce{NO2} concentrations in Madrid 2023. Elevated hourly averages correspond with rush-hour traffic in the morning and in the evening. Spatial detail is resolved to high-resolution with distinct road-network pattern of high \ce{NO2} levels, and markedly lower levels in green-spaces.}
    \label{fig:madrid_hourly}
\end{figure}

The model covers the different temporal dynamics of \ce{NO2} well and accurately represents seasonal cycles with highs in the Winter, daily cycles around rush-hour in the morning and evenings (Fig.~\ref{fig:madrid_hourly}), as well as reduced emissions on the weekend relative to the rest of the week (Fig.~\ref{fig:temporal_cycles}). On the annual scale, the model follows the general downward-trend of \ce{NO2} during the 2018-2023 period (Fig.~\ref{fig:by_year}). As baselines, we include a constant mean prediction, which yields the highest RMSE in our evaluation, as well as an evaluation of CAMS \ce{NO2} reanalysis data alone, which already possesses significant skill (R2-score up to $0.41$) despite significantly lower spatial resolution. To isolate the effect of spatial context through the multi-modal input patches ($1.28\times 1.28$\,km) we compare the factored U-net ensemble model with a ridge regression~\cite{horel1962application} baseline trained on the center pixels of the full multi-modal input patches. This yields slightly lower \ce{NO2} estimation accuracy than the factored U-net ensemble (Tab.~\ref{tab:results_heldout_monitors}), which leverages the full spatial context (R2-score $0.517\pm0.011$ vs. $0.571\pm0.013$).

We investigate the predictive performance of the factored U-net ensemble model across different bins of observed \ce{NO2} concentration (Tab.~\ref{tab:v13-bias-by-bin}). More than half of the station-hours of the held-out monitors are below 20\,\ugm{} (16.7\,M samples). In this range, the model exhibits slightly positive bias and low errors (RMSE $6.11$\,\ugm{} in the $[0-10]$ bin). With increasing observed \ce{NO2} concentration, the model bias skews increasingly negative. The predictive uncertainty also increases, although at a lower rate. Approximately \num{6.1}\% of the held-out station-hours are above the 50\,\ugm{} threshold, making them relevant for the EU daily \ce{NO2} limit. For these observations, the model has an average bias of \num{-24.75}\,\ugm{}, with a predictive $\bar\sigma$ of \num{16.16}.
\begin{table}[t]
\centering
\caption{\textbf{Accuracy and predictive uncertainty by observed \ce{NO2} bin.} Across all 25\,M station-hours of the held-out monitors.  $\bar{\sigma}$ is the mean predictive standard deviation of the factored UNet ensemble. Coverage is the share of hours inside the nominal 95\,\% predictive interval. Bias, MAE, RMSE and $\bar{\sigma}$ are in \si{\micro\gram\per\cubic\metre}.}
\label{tab:v13-bias-by-bin}
\small
\begin{tabular}{@{}lrrrrrrr@{}}
\toprule
Observed bin & Num. samples & Frac. samples & Bias & MAE & RMSE & $\bar{\sigma}$ & Cov.\ 95\% \\
\midrule
0--10 & \num{10355701} & 41.5 & +3.71 & 4.18 & 6.11 & 4.52 & 98.6 \\
10--20 & \num{6346747} & 25.5 & +2.56 & 5.52 & 7.57 & 7.81 & 95.5 \\
20--30 & \num{3447764} & 13.8 & $-$1.09 & 7.22 & 9.14 & 10.18 & 90.3 \\
30--40 & \num{2048863} & 8.2 & $-$5.74 & 9.97 & 12.24 & 11.95 & 84.1 \\
40--50 & \num{1207531} & 4.8 & $-$11.08 & 13.81 & 16.68 & 13.43 & 77.5 \\
50--75 & \num{1194166} & 4.8 & $-$20.04 & 21.41 & 25.19 & 15.42 & 66.9 \\
75--100 & \num{255245} & 1.0 & $-$36.68 & 37.07 & 41.15 & 18.29 & 48.5 \\
$\geq$100 & \num{67528} & 0.3 & $-$63.03 & 63.13 & 69.45 & 21.30 & 28.7 \\
\midrule
$\geq$50 & \num{1516939} & 6.1 & $-$24.75 & 25.90 & 31.61 & 16.16 & 62.1 \\
\bottomrule
\end{tabular}
\end{table}

\textbf{Passive samplers.}
\begin{table}[tb]
\centering
\caption{\textbf{Citizen science campaigns.} Evaluation of the factored U-net \ce{NO2} estimates against independent passive-sampler campaigns.}
\label{tab:passive-campaign-v13}
\small
\begin{tabular}{@{}lrrrrrrrr@{}}
\toprule
& & \multicolumn{2}{c}{Mean} & & & & & \\
\cmidrule(lr){3-4}
Campaign & $n$ & obs. & pred. & Bias & RMSE & MAE & $r$ & $R^2$ \\
\midrule
Brussels, 2021$^{a}$   & 2\,493  & 24.43 & 23.80 & $-0.63$ & 5.05 & 3.85 & 0.697 & 0.470 \\
Flanders, 2018$^{b}$ & 17\,886 & 21.33 & 20.25 & $-1.09$ & 4.25 & 2.99 & 0.772 & 0.562 \\
\bottomrule
\end{tabular}

\vspace{4pt}
{\footnotesize\raggedright
$^{a}$~Each tube averaged over its own exposure window (16--61\,d, median 28\,d, 25\,Sep--23\,Oct 2021) and compared with the campaign's measurement-period mean.\par
$^{b}$~Full calendar month May~2018 for every tube.\par
}
\end{table}
We evaluate model predictions at the independent citizen science campaigns with passive \ce{NO2} samplers, aggregated to each campaign's exposure window. We find strong predictive skill for the Brussels (\num{2493} samplers) and Flanders-wide (\num{17886} samplers) campaigns (RMSE \num{5.05} and \num{4.25}\,\ugm{}, respectively), accurately capturing spatial variance in densely monitored regions.

\textbf{Modality selection.}
In order to achieve the best possible skill at high-resolution \ce{NO2} estimation, we collect a comprehensive set of input modalities. We reason that the factored U-net ensemble handles potential collinearity between the predictors well, and learns to disregard uninformative modalities. We test this in a leave-one-out ablation study, where the model is re-trained on subsets of the input modalities (Tab.~\ref{tab:abl-modality}). We find that, even in the presence of all other modalities, some individual modalities are extraordinarily important for \ce{NO2} estimation performance. The AlphaEarth embeddings are the most important static input, while CAMS is the most important dynamic modality. Only the GRIP4 road network and GHSL population data have a significant negative effect on overall performance. However, both of these modalities have direct analogues in the remaining dataset (WorldPop and the street network rasterized from OSM).
\begin{table}[t]
\centering
\caption{\textbf{Leave-one-out ablation of input modalities.} We remove input modalities one at a time and retrain the factored u-Net model with all other settings constant. Each is a 3-seed ensemble scored on the same held-out station-hours at 636 stations. $\Delta$RMSE is computed from paired bootstrap resamples and defined as RMSE(ablated)~$-$~RMSE(production). Intervals are 95\,\% station-clustered bootstrap (1000 resamples). $^{*}$ marks an interval excluding zero.}
\label{tab:abl-modality}
\small
\begin{tabular}{@{}lrrl@{}}
\toprule
Modality removed & Channels & RMSE & $\Delta$RMSE \\
\midrule
None (full inputs) & 163 & 11.40 & --- \\
\midrule
\multicolumn{4}{@{}l}{\emph{Static modalities}} \\
\addlinespace[2pt]
AEF land-use embedding & 64 & 11.60 & +0.20 $\pm$ 0.05$^{*}$ \\
WorldPop density & 1 & 11.45 & +0.05 $\pm$ 0.04 \\
CAMS-REG traffic & 16 & 11.45 & +0.05 $\pm$ 0.05 \\
AADT traffic volume & 10 & 11.44 & +0.05 $\pm$ 0.05 \\
Lat/lon + year & 3 & 11.44 & +0.05 $\pm$ 0.05 \\
OpenStreetMap & 16 & 11.43 & +0.04 $\pm$ 0.05 \\
ESA WorldCover & 11 & 11.41 & +0.01 $\pm$ 0.03 \\
DEM + wavelets & 13 & 11.38 & $-$0.02 $\pm$ 0.06 \\
GHSL pop. + GRIP4 road & 3 & 11.27 & $-$0.13 $\pm$ 0.06$^{*}$ \\
\midrule
\multicolumn{4}{@{}l}{\emph{Dynamic modalities}} \\
\addlinespace[2pt]
CAMS regional \ce{NO2} & 1 & 12.47 & +1.08 $\pm$ 0.10$^{*}$ \\
Time encodings & 8 & 11.68 & +0.29 $\pm$ 0.05$^{*}$ \\
ERA5 meteorology (+hdd) & 7 & 11.49 & +0.09 $\pm$ 0.05 \\
S5P column (+missing mask) & 4 & 11.31 & $-$0.09 $\pm$ 0.05 \\
CAMS-GLOB-ANT \ce{NO_x} emissions & 6 & 11.31 & $-$0.09 $\pm$ 0.05 \\
\bottomrule
\end{tabular}
\end{table}

\textbf{Architecture and capacity.}
\begin{table}[t]
\centering
\caption{\textbf{Model capacity.} The width of the factored UNet stem and encoder is varied to produce different capacity variants, with all other settings held constant. Each model is a 3-seed ensemble scored on the same held-out station-hours at 636 stations. $\Delta$RMSE is paired bootstrap resamples and intervals are 95\,\% station-clustered bootstrap (1000 resamples)}.
\label{tab:abl-capacity}
\small
\begin{tabular}{@{}lrrrr@{}}
\toprule
Parameters (M) & Encoder & Stem & RMSE & $\Delta$RMSE \\
\midrule
0.61 & 32 & 64 & 11.44 & +0.05 $\pm$ 0.05 \\
2.25 & 64 & 128 & \textbf{11.40} & --- \\
4.93 & 96 & 192 & 11.42 & +0.02 $\pm$ 0.03 \\
8.65 & 128 & 256 & 11.45 & +0.04 $\pm$ 0.05 \\
\bottomrule
\end{tabular}
\end{table}

We instantiate the factored U-net model with different sizes and parameter counts by varying the model stem (a convolutional layer that maps the input stack channels to the U-net encoder size) and encoder widths (Tab.~\ref{tab:abl-capacity}). The model is varied across a $13$--$14\times$ parameter span, but the model size has little effect on the performance. We adopt the variant with stem and encoder sizes of 64 and 128, respectively, for this work, resulting in a \num{2.25}M parameter model.

\begin{table}[t]
\centering
\caption{\textbf{Temporal factorization rank.} The number of temporal modes $K$ is varied, with all other settings held constant. Each model is a 3-seed ensemble scored on the same held-out station-hours at 636 stations. $\Delta$RMSE is paired bootstrap resamples and intervals are 95\,\% station-clustered bootstrap (1000 resamples).}
\label{tab:abl-temporal-rank}
\small
\begin{tabular}{@{}lrr@{}}
\toprule
Temporal rank $K$ & RMSE & $\Delta$RMSE \\
\midrule
2 & 11.44 & +0.04 $\pm$ 0.04 \\
4 & \textbf{11.40} & --- \\
8 & 11.41 & +0.02 $\pm$ 0.04 \\
\bottomrule
\end{tabular}
\end{table}

To find the ideal number of temporal mode maps $K$ for the factored U-net we ablate $K$ in the range of $2-8$ and find $K=3$ to be optimal, but only slightly ahead in RMSE of other configurations (Tab.~\ref{tab:abl-temporal-rank}).

\begin{table}[t]
\centering
\caption{\textbf{Deep-ensemble size.} Performance of factored UNet ensembles by ensemble size. We draw all possible subsets from the full 5-member ensemble and report average performance metrics per subset size.}
\label{tab:abl-ensemble}
\small
\begin{tabular}{@{}lrrrr@{}}
\toprule
Members & Subsets & RMSE & $R^2$ & $\Delta$RMSE vs.\ 5 \\
\midrule
1 & 5 & 11.492 & 0.5581 & +0.167 $\pm$ 0.040 \\
2 & 10 & 11.388 & 0.5661 & +0.063 $\pm$ 0.029 \\
3 & 10 & 11.353 & 0.5688 & +0.028 $\pm$ 0.019 \\
4 & 5 & 11.335 & 0.5701 & +0.011 $\pm$ 0.013 \\
5 & 1 & \textbf{11.325} & 0.5709 & --- \\
\bottomrule
\end{tabular}
\end{table}

We investigate the benefits of increasing the ensemble size relative to the full, 5-member ensemble used throughout this work. For $1-5$ ensemble members, we select all possible subsets of that cardinality from the full ensemble and compute the average performance of ensembles formed from each subsets. We find significant, although diminishing marginal, improvement in RMSE and R2-score with increasing ensemble size (Tab.~\ref{tab:abl-ensemble}).

Hyperparameters are likewise only modestly influential. A Bayesian hyperband~\cite{li2018hyperband} search over learning rate, weight decay, batch size and oversampling improved RMSE by $\approx0.15$. We find that, by a slight margin, the following hyperparameter configuration is ideal, and adapt it as the standard setting: learning rate $5.3\times10^{-4}$ region, weight decay $7.9\times10^{-5}$, batch size 16, and oversampling of high-concentration locations with $\alpha=0.25$.

\begin{figure}
    \centering
    \includegraphics[width=1.\linewidth]{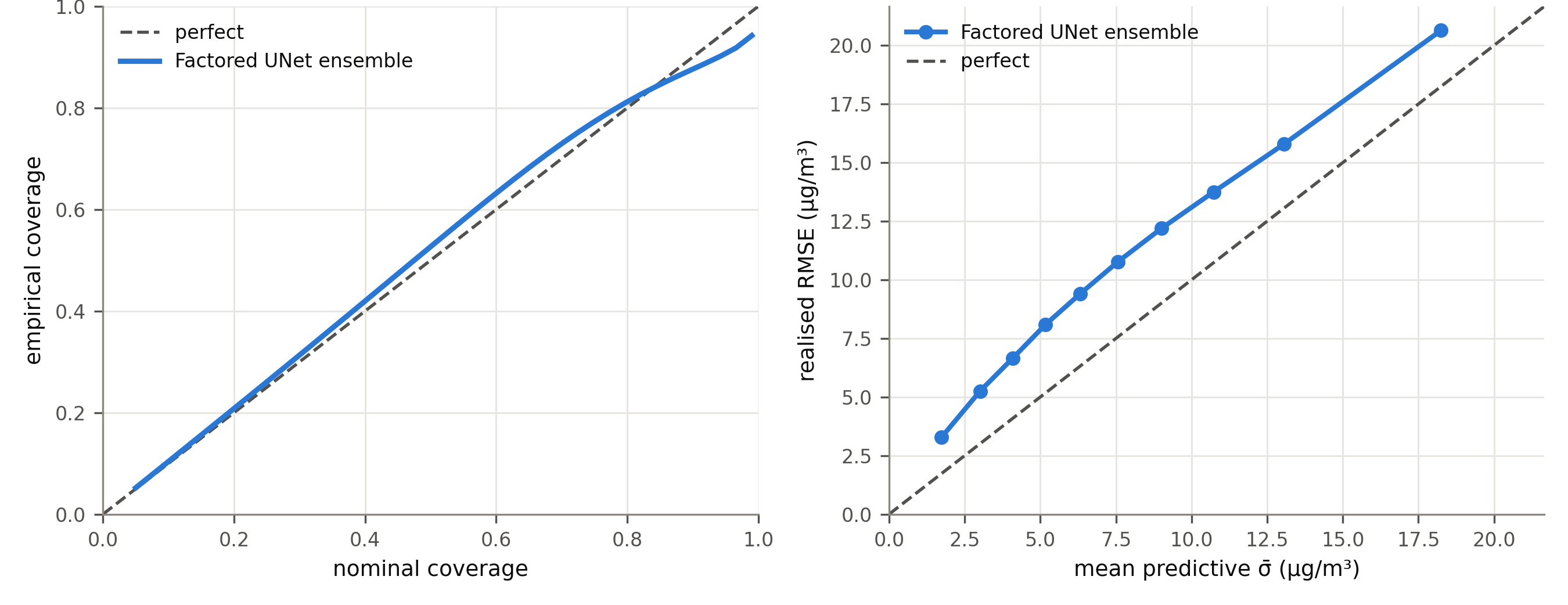}
    \caption{Calibration of the factored U-net ensemble's predictive uncertainty. \textbf{Left:} The expected calibration error is 0.020 with empirical coverage closely aligned to nominal. \textbf{Right:} Realized RMSE binned by decile of predictive $\bar\sigma$. The model's uncertainty consistently rises with true error magnitude, although the realized error consistently exceeds predictive uncertainty.}
    \label{fig:uncertainty_calibration}
\end{figure}

\textbf{Uncertainty.}
We find that the factored U-net ensemble's predictive $N(\bar\mu,\bar\sigma^2)$ distribution is well calibrated (Fig.~\ref{fig:uncertainty_calibration}). The calibration ratio (RMSE/RMS $\bar\sigma$) is 1.26, expected calibration error 0.020, and the nominal 68\%/95\% predictive intervals achieve empirical coverage of 71\%/91\%, indicating some overconfidence. The predictive uncertainty $\bar\sigma$ is an indicator of genuine difficulty. When binning station-hours into deciles of predictive $\bar\sigma$, realized RMSE rises monotonically from \num{3.3} to \num{20.6}\,\ugm{} across the decile range, and the row-level correlation between absolute error and $\bar\sigma$ is 0.43. Calibration is not uniform across station strata, with traffic (ratio 1.34) sites more overconfident than background (1.09) and rural (1.06) sites. However, calibration is approximately stable across years (1.18–1.39).

\textbf{Breach detection.}
\begin{table}[t]
\centering
\caption{\textbf{Non-compliance detection.} Limit breach detection capabilities of CAMS \ce{NO2} reanalysis, exceed count from the factored U-net, and the probabilistic breach statistics. CAMS makes no detections while all model-derived metrics have high precision but underestimate the total number of limit breaches. CAMS is calculated across \num{3060} regulatory monitors in 2023, model-based metrics are reported on 2488 held-out location-years.}
\label{tab:breach-confusion}
\small
\begin{tabular}{@{}lrrrrrrr@{}}
\toprule
Statistic & TP & FP & FN & TN & Precision & Recall & F1 \\
\midrule
CAMS exceed count (2023) & 0 & 0 & 264 & 2{,}796 & 0 & 0 & 0\\
Model exceed count & 101 & 8 & 257 & 2{,}122 & 0.927 & 0.282 & 0.433 \\
Expected number of exceeding days & 124 & 15 & 234 & 2{,}115 & 0.892 & 0.346 & 0.499 \\
Probability of non-compliance & 122 & 12 & 236 & 2{,}118 & 0.910 & 0.341 & 0.496 \\
\bottomrule
\end{tabular}
\end{table}

\begin{figure}
\centering
\begin{subfigure}{.5\textwidth}
  \centering
  \includegraphics[width=1.\linewidth]{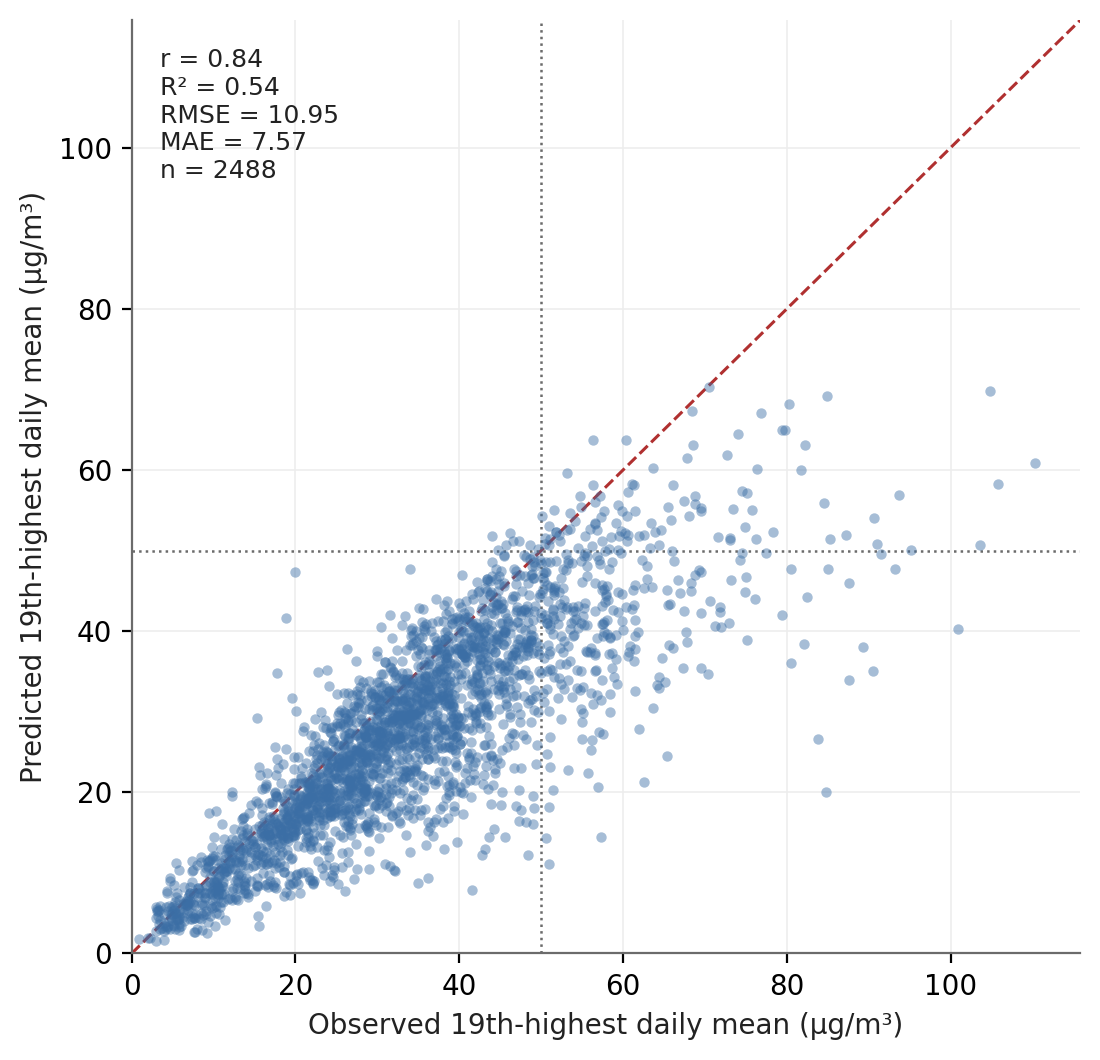}
\end{subfigure}%
\begin{subfigure}{.5\textwidth}
  \centering
  \includegraphics[width=1.\linewidth]{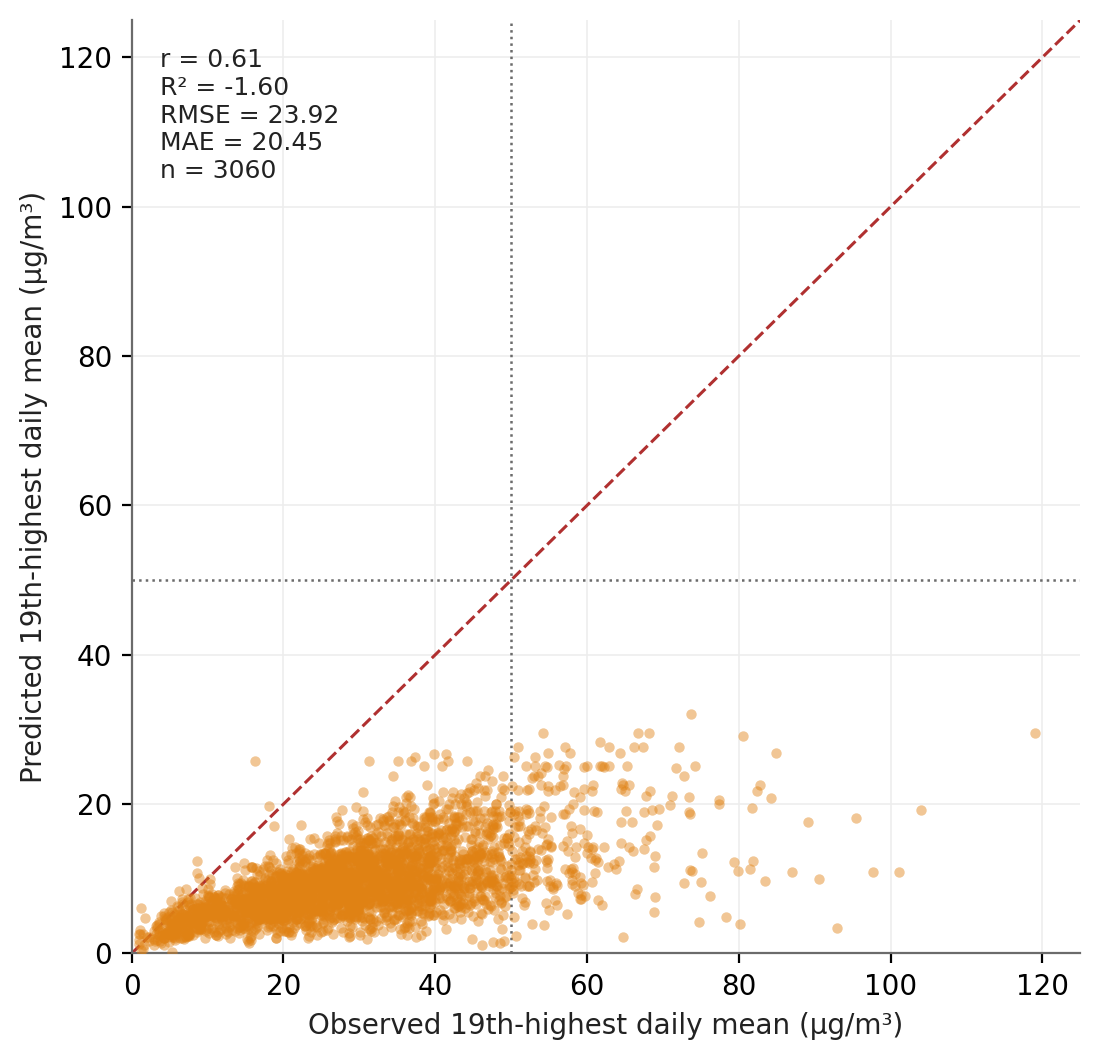}
\end{subfigure}
\caption{\textbf{Average \ce{NO2} on the 19\textsuperscript{th} highest day of the year.} Observed versus predicted average daily \ce{NO2} concentration for the factored U-net (left, \num{2488} held-out locations) and CAMS models (right, \num{3060} locations in  2023). Dashed lines indicate the 50\,\ugm{} limit.}
\label{fig:observed_vs_predicted_19th}
\end{figure}
We quantify the quality the proposed expected and probabilistic breach statistics on the held-out regulatory monitors (Tab.~\ref{tab:breach-confusion}). Both statistics exhibit high precision ($0.892$ and $0.927$, respectively). This indicates that detected limit breaches correspond to true breaches with high likelihood (Fig.~\ref{fig:observed_vs_predicted_19th}, left). However, the statistics also underestimate the total number of breaches. Across 2488 station-years, our hold-out dataset contains 358 true annual limit breaches, $234-236$, of which are not detected (recall $0.341-0.346$). The reported number of breaches is therefore a conservative lower bound on the total number of breaching \SI{10}{\metre}\textsuperscript{2} locations in the mapped regions. 
We compare the probabilistic metrics with the deterministic count of days above the exposure limit computed from the modeled hourly timeseries ($\sum_{d=1}^D\mathds{1}[\mu_d>50]$). This yields similar aggregate metrics to the other compliance statistics (precision $0.927$, recall $0.282$), but lacks the probabilistic interpretation.

Additionally, we compute limit-breach statistics from CAMS data for the year 2023. Across all \num{3060} regulatory \ce{NO2} monitors, 264 limit breaches were recorded. The CAMS \ce{NO2} reanalysis with \SI{10}\,km GSD misses all of the true breaches, estimating limit breaching conditions at none of the regulatory monitor locations (Tab.~\ref{tab:breach-confusion}, Fig.~\ref{fig:observed_vs_predicted_19th}, right). We interpret this as an artifact of averaging highly-varying local \ce{NO2} levels across the \SI{10}\,km grid, motivating high-resolution \ce{NO2} modeling.   

\subsection{Air quality zones}
Compliance with the EU \ce{NO2} 24-hour limit is assessed and reported per air-quality zone~\cite{EU2024_2881}. We obtain zone delineations and their reported resident populations from the EEA e-Reporting portal~\cite{eea_air_quality_reporting}. For each country, we select air quality zones designated for health monitoring and \ce{NO2} for 2023. Any duplicate zones are removed, and overlaps are resolved by selecting the smallest applicable zone. This results in \num{670} zones across Europe (excluding Switzerland, San Marino, Andorra), with a resident population of \num{520.8}\,M.
To identify breaching zones from modeled hourly \ce{NO2} concentrations we form daily means from the full-year hourly predictive distribution. We then use the probabilistic compliance statistics to flag any pixels that are estimated to be in breach of the limit. Following the regulatory monitoring convention, a zone is deemed in breach where at least one mapped pixel is flagged.  Throughout our analyses, zones are generally mapped only in part, hence the absence of a modeled limit breach only indicates that no exceedance was found within the mapped fraction of the zone, not that the zone as a whole is compliant.

\subsection{Population exposure}
We translated the predicted exceedance fields into the population living under a predicted breach of the EU 24-hour \ce{NO2} limit value. For each of the 110 mapped regions we derived a binary breach mask at \SI{10}{\metre} resolution by thresholding the non-compliance probability and combine it with gridded residential population from the Global Human Settlement layer (GHS-POP R2023A)~\cite{schiavina2023ghs}. Each \SI{100}{\metre} population cell contributes its count weighted by the fraction of its area under a predicted breach. Alongside this deterministic estimate we compute an uncertainty-aware count as the expected affected population from the product of the population count and the non-compliance probability. This approach requires no threshold and integrates the model's per-pixel exceedance probability.

\subsection{Breaches on roads}\label{sec:road_breaches}
Exhaust fumes from vehicles powered by internal combustion engines are one of the main anthropogenic sources of \ce{NO2}~\cite{grice2009recent}, and \ce{NO2} levels are commonly elevated on roads. However, locations at the center of roads or busy intersections are often not representative of the conditions even in the near vicinity. We therefore conduct an analysis where we exclude limit breaches detected on roads explicitly from the obtained results. We obtain road network data from OpenStreetMap~\cite{OpenStreetMap} and assume road widths at two different levels, \emph{narrow} and \emph{wide} (Tab.~\ref{tab:road_widths}), for the road vectors. For this analysis, any limit breaching pixel that intersects with the rasterized road data is disregarded.

\begin{table}[]
    \centering
    \begin{tabular}{c|c|c|c|c|c}
    \toprule
    OSM class & motorway, trunk & primary & secondary & tertiary & rest  \\
     \midrule
    narrow & $\pm$\SI{10}{\metre} & $\pm$\SI{6}{\metre} & $\pm$\SI{5}{\metre} & $\pm$\SI{4}{\metre} & $\pm$\SI{3}{\metre} \\
    wide & $\pm$\SI{20}{\metre} & $\pm$\SI{12}{\metre} & $\pm$\SI{10}{\metre} & $\pm$\SI{8}{\metre} & $\pm$\SI{6}{\metre} \\
    \bottomrule
    \end{tabular}
    \caption{\textbf{Road widths.} Road widths attributed to different OSM road classes under \emph{narrow} and \emph{wide} assumptions. The buffer is applied to both sides of the street centerline.}
    \label{tab:road_widths}
\end{table}



\section{Data Availability}


\bigskip\noindent
The compliance statistics can be downloaded at \url{https://doi.org/10.6084/m9.figshare.33257289} and viewed at \url{https://scheibenreif.users.earthengine.app/view/no2compliancestatistics}.

\bigskip\noindent
AlphaEarth foundations embeddings are available at: \url{https://source.coop/tge-labs/aef}

\bigskip\noindent
Copernicus GLO-30 digital elevation model is available at: \url{https://dataspace.copernicus.eu/explore-data/data-collections/copernicus-contributing-missions/collections-description/COP-DEM}

\bigskip\noindent
OpenStreetMap is available from: \url{https://openstreetmap.org}

\bigskip\noindent
WorldPop population data is available from: \url{https://worldpop.org}

\bigskip\noindent
CAMS data is available at: \url{https://atmosphere.copernicus.eu/}

\bigskip\noindent
ESA WorldCover is available at: \url{https://esa-worldcover.org/en}

\bigskip\noindent
ERA5 data is available at: \url{https://cds.climate.copernicus.eu/datasets}

\bigskip\noindent
Sentinel-5P data is available at: \url{https://dataspace.copernicus.eu/data-collections/copernicus-sentinel-missions/sentinel-5p}

\bigskip\noindent
GHSL is available at: \url{https://human-settlement.emergency.copernicus.eu/}

\bigskip\noindent
AADT traffic flow estimates are available via: \url{https://github.com/co822ee/eu_roadTraffic}

\bigskip\noindent
EEA air pollution monitoring data: \url{https://www.eea.europa.eu/en/datahub}

\section{Code Availability}
The code to analyze the \ce{NO2} maps is available at \url{https://github.com/scheibenreif/no2_exposure_europe}.

\backmatter

\bmhead{Acknowledgements}

We thank Prof.\,Dr.\,Dominik Brunner and the Atmospheric Modeling group at Empa for helpful discussions. We would also like to thank Prof.\,Dr.\,Filip Meysman at the University of Antwerp for making the citizen science \ce{NO2} data from Flanders and Brussels available to us.









\begin{appendices}

\section{Mapped regions}\label{secA1}
\begin{table*}[t]
\centering
\caption{Dense 10\,m NO$_2$ compliance-mapped regions (110 cities and regions. Grid gives the raster dimensions in \SI{10}{\metre} pixels, area is the mapped land extent.}
\label{tab:mapped_regions}
\footnotesize
\setlength{\tabcolsep}{4pt}
\begin{minipage}[t]{0.49\linewidth}\centering
\begin{tabular}{lrr}
\toprule
Region & Grid (px) & Area (km$^2$) \\
\midrule
Po Valley & 11423$\times$22255 & 24,287 \\
Midlands North & 18123$\times$13057 & 23,153 \\
Flanders & 9919$\times$24001 & 20,516 \\
Rhine Ruhr & 11780$\times$12776 & 14,625 \\
Upper Silesia & 7729$\times$12460 & 9,336 \\
Paris & 9178$\times$9167 & 8,246 \\
Randstad & 8767$\times$8272 & 5,413 \\
Madrid & 7031$\times$7043 & 4,862 \\
Rome & 6183$\times$6195 & 3,296 \\
Berlin & 5699$\times$5702 & 3,086 \\
London & 5080$\times$5844 & 2,837 \\
Munich & 5389$\times$5411 & 2,795 \\
Hamburg & 5386$\times$5409 & 2,778 \\
Nottingham & 5339$\times$5361 & 2,776 \\
Geneva & 5486$\times$5494 & 2,706 \\
Glasgow & 5258$\times$5259 & 2,624 \\
Vienna & 5184$\times$5204 & 2,609 \\
Warsaw & 5097$\times$5167 & 2,552 \\
Belgrade & 5097$\times$5108 & 2,483 \\
Frankfurt & 5020$\times$5032 & 2,464 \\
Oslo & 5350$\times$5370 & 2,416 \\
Zurich & 5088$\times$5097 & 2,415 \\
Belfast & 5568$\times$5569 & 2,412 \\
Prague & 4929$\times$4936 & 2,367 \\
Southampton & 5284$\times$5305 & 2,340 \\
Stuttgart & 4850$\times$4895 & 2,333 \\
Bristol & 5074$\times$5094 & 2,290 \\
Lyon & 4832$\times$4851 & 2,260 \\
Barcelona & 6516$\times$6526 & 2,204 \\
Rotterdam & 5084$\times$5104 & 2,146 \\
Mannheim & 4712$\times$4721 & 2,145 \\
Athens & 5523$\times$5549 & 2,133 \\
Bucharest & 4611$\times$4623 & 2,052 \\
Stockholm & 5194$\times$5195 & 2,029 \\
Newcastle & 5263$\times$5283 & 2,021 \\
Budapest & 4471$\times$4483 & 1,929 \\
Edinburgh & 5056$\times$5098 & 1,916 \\
Turin & 4358$\times$4372 & 1,857 \\
Bologna & 4336$\times$4349 & 1,852 \\
Seville & 4276$\times$4290 & 1,782 \\
Florence & 4224$\times$4237 & 1,781 \\
Padua & 4228$\times$4239 & 1,775 \\
Liverpool & 5020$\times$5091 & 1,775 \\
Lisbon & 4971$\times$5009 & 1,755 \\
Naples & 5192$\times$5205 & 1,708 \\
Brighton & 5507$\times$5529 & 1,637 \\
Valencia & 5052$\times$5075 & 1,621 \\
Nuremberg & 4029$\times$4040 & 1,619 \\
Toulouse & 4025$\times$4038 & 1,608 \\
Luxembourg & 3979$\times$3986 & 1,581 \\
Wroclaw & 3980$\times$3989 & 1,572 \\
Pecs & 3940$\times$3951 & 1,550 \\
Arnhem & 3999$\times$4008 & 1,542 \\
Lille & 3916$\times$3958 & 1,542 \\
Kaunas & 4022$\times$4029 & 1,540 \\
\bottomrule
\end{tabular}
\end{minipage}\hfill
\begin{minipage}[t]{0.49\linewidth}\centering
\begin{tabular}{lrr}
\toprule
Region & Grid (px) & Area (km$^2$) \\
\midrule
Grenoble & 3925$\times$3935 & 1,532 \\
Bolzano & 3896$\times$3908 & 1,519 \\
Leipzig & 3952$\times$3961 & 1,513 \\
Bordeaux & 3945$\times$3955 & 1,509 \\
Plovdiv & 3843$\times$3855 & 1,474 \\
Poznan & 3852$\times$3860 & 1,463 \\
Zaragoza & 3821$\times$3833 & 1,456 \\
Hannover & 3831$\times$3841 & 1,455 \\
Vilnius & 3829$\times$3837 & 1,450 \\
Bratislava & 3857$\times$3868 & 1,438 \\
Rouen & 3834$\times$3843 & 1,438 \\
Bielefeld & 3791$\times$3800 & 1,436 \\
Lodz & 3790$\times$3799 & 1,436 \\
Avignon & 3810$\times$3822 & 1,429 \\
Murcia & 3770$\times$3785 & 1,419 \\
Munster & 3765$\times$3773 & 1,415 \\
Pamplona & 3729$\times$3741 & 1,391 \\
Dresden & 3745$\times$3755 & 1,390 \\
Nantes & 3760$\times$3770 & 1,378 \\
Strasbourg & 3736$\times$3747 & 1,355 \\
Zagreb & 3690$\times$3700 & 1,354 \\
Ploiesti & 3687$\times$3699 & 1,353 \\
Klagenfurt & 3706$\times$3717 & 1,350 \\
Wiesbaden & 3669$\times$3679 & 1,326 \\
Benevento & 3631$\times$3644 & 1,320 \\
Bremen & 3651$\times$3670 & 1,316 \\
Ljubljana & 3625$\times$3635 & 1,313 \\
Granada & 3620$\times$3635 & 1,312 \\
Graz & 3619$\times$3631 & 1,309 \\
Dublin & 4264$\times$4271 & 1,288 \\
Timisoara & 3584$\times$3600 & 1,287 \\
Riga & 4028$\times$4034 & 1,254 \\
Piombino & 3782$\times$3793 & 1,242 \\
Gothenburg & 4046$\times$4052 & 1,199 \\
Algeciras & 3832$\times$3846 & 1,194 \\
Thessaloniki & 3793$\times$3808 & 1,189 \\
Bilbao & 3565$\times$3597 & 1,183 \\
Kiel & 3738$\times$3747 & 1,130 \\
Gdansk & 3926$\times$3934 & 1,115 \\
Porto & 4126$\times$4141 & 1,103 \\
Marseille & 4328$\times$4521 & 1,099 \\
Sofia & 3055$\times$3516 & 1,066 \\
Montpellier & 3670$\times$3682 & 1,054 \\
Perpignan & 3568$\times$3595 & 1,008 \\
Copenhagen & 4766$\times$4772 & 973 \\
Tallinn & 3934$\times$3941 & 944 \\
Malaga & 3712$\times$3727 & 926 \\
Laspezia & 3673$\times$3684 & 883 \\
Helsinki & 4282$\times$4289 & 882 \\
Nice & 3783$\times$3793 & 833 \\
Varna & 3674$\times$3686 & 827 \\
Palermo & 3743$\times$3758 & 807 \\
Bari & 3788$\times$3801 & 750 \\
Catania & 3564$\times$3591 & 742 \\
Genoa & 3567$\times$3598 & 690 \\
\bottomrule
\end{tabular}
\end{minipage}
\end{table*}
\begin{figure}
    \centering
    \includegraphics[width=1\linewidth]{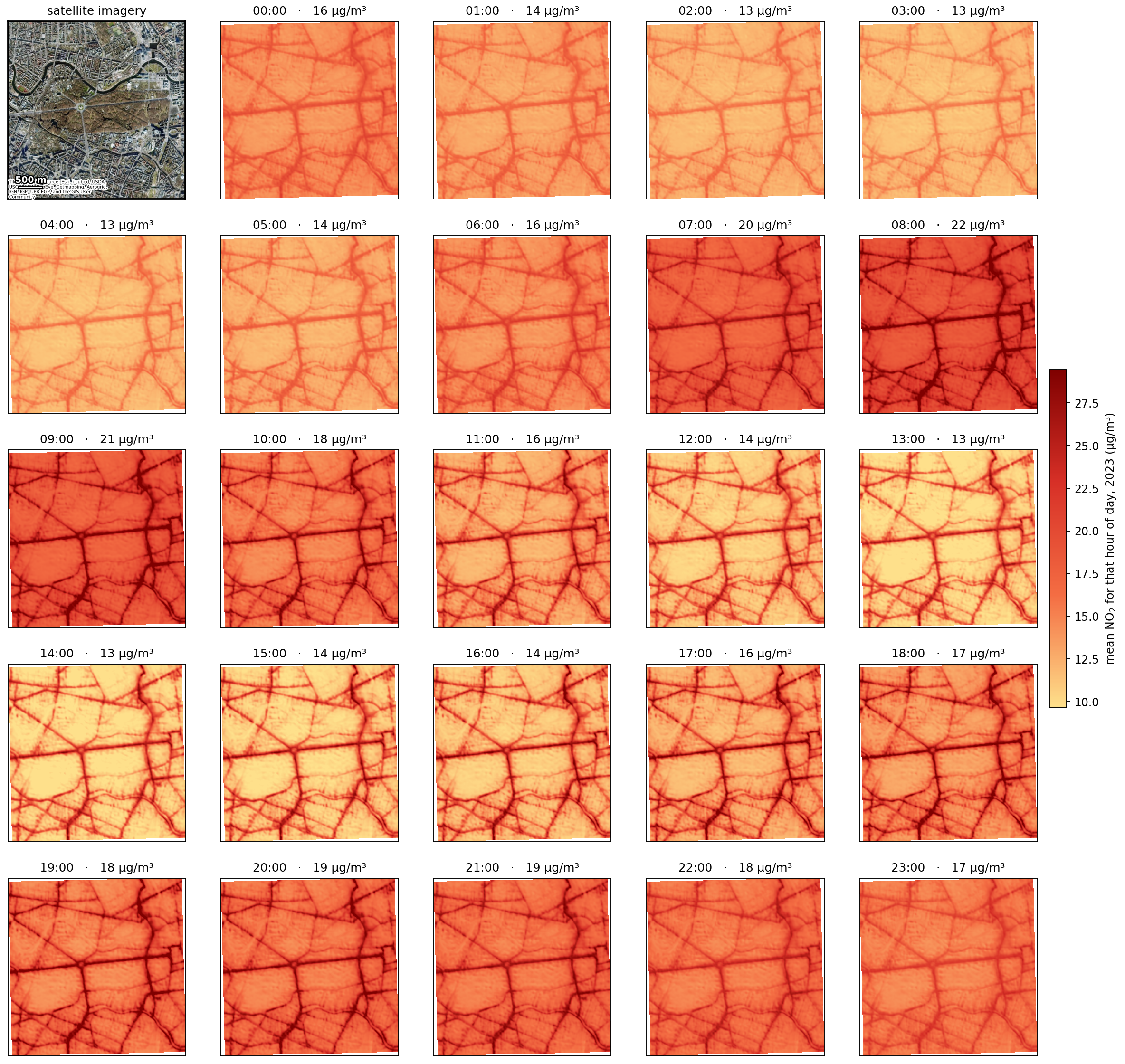}
    \caption{\textbf{Hourly \ce{NO2} maps for Berlin.}}
    \label{fig:placeholder}
\end{figure}



\end{appendices}


\bibliographystyle{unsrt}
\bibliography{sn-bibliography}

\end{document}